%% file: main.tex
\documentclass[10pt,twocolumn,letterpaper]{article}

\usepackage[pagenumbers]{cvpr} 
\usepackage[T1]{fontenc}
\usepackage[utf8]{inputenc}
\usepackage[table]{xcolor}
\usepackage{tabularx}
\usepackage{booktabs}
\usepackage{multirow}
\usepackage{array}
\usepackage{stfloats}

\input{preamble}
\definecolor{cvprblue}{rgb}{0.21,0.49,0.74}
\usepackage[pagebackref,breaklinks,colorlinks,allcolors=cvprblue]{hyperref}

\def\paperID{} 
\def\confName{CVPR}
\def\confYear{2026}

\newcommand{\KalphaLOS}{K\texorpdfstring{$\alpha$}{α}LOS}
\newcommand{\KripA}{K-\texorpdfstring{$\alpha$}{α}}
\newcommand{\Deltaloc}{\texorpdfstring{$\Delta_{loc}$}{Δ\_{loc}}}
\newcommand{\Deltacls}{\texorpdfstring{$\Delta_{cls}$}{Δ\_{cls}}}
\newcommand{\Deltaum}{\texorpdfstring{$\Delta_{um}$}{Δ\_{um}}}
\newcommand{\Deltatopo}{\texorpdfstring{$\Delta_{topo}$}{Δ\_{topo}}}

\title{\KalphaLOS{} finds Consensus: A Meta-Algorithm for Evaluating Inter-Annotator Agreement in Complex Vision Tasks}

\author{
    David Tschirschwitz \quad Volker Rodehorst\\ \\
    Bauhaus-Universität Weimar, Germany\\
    {\tt\small david.tschirschwitz@uni-weimar.de} 
}

\begin{document}
\maketitle
\input{sec/0_abstract}    
\input{sec/1_intro}
\input{sec/2_related_work}
\input{sec/3_kalos}
\input{sec/4_experimental_validation_framework}
\input{sec/5_experiment_and_results}
\input{sec/6_conclusion}
{
    \small
    \bibliographystyle{ieeenat_fullname}
    \bibliography{main}
}

\input{sec/7_appendix}

\end{document}

%% file: sec/0_abstract.tex
\begin{abstract}

Progress in object detection benchmarks is stagnating. It is limited not by architectures but by the inability to distinguish model improvements from label noise. To restore trust in benchmarking the field requires rigorous quantification of annotation consistency to ensure the reliability of evaluation data. However, standard statistical metrics fail to handle the instance correspondence problem inherent to vision tasks. Furthermore, validating new agreement metrics remains circular because no objective ground truth for agreement exists. This forces reliance on unverifiable heuristics. 

We propose \KalphaLOS{}, a unified meta-algorithm that generalizes the "Localization First" principle to standardize dataset quality evaluation. By resolving spatial correspondence before assessing agreement, our framework transforms complex spatio-categorical problems into nominal reliability matrices. Unlike prior heuristic implementations, \KalphaLOS{} employs a principled, data-driven configuration; by statistically calibrating the localization parameters to the inherent agreement distribution, it generalizes to diverse tasks ranging from bounding boxes to volumetric segmentation or pose estimation. This standardization enables granular diagnostics beyond a single score. These include annotator vitality, collaboration clustering, and localization sensitivity. To validate this approach, we introduce a novel and empirically derived noise generator. Where prior validations relied on uniform error assumptions, our controllable testbed models complex and non-isotropic human variability. This provides evidence of the metric’s properties and establishes \KalphaLOS{} as a robust standard for distinguishing signal from noise in modern computer vision benchmarks.

Code for \KalphaLOS{}: \href{https://github.com/Madave94/kalos}{https://github.com/Madave94/kalos}
\end{abstract}

\begin{figure}[!t]
    \centering
    
    \begin{subfigure}[t]{0.48\linewidth}
        \centering
        \includegraphics[width=\linewidth,trim={0 0 0 0},clip]{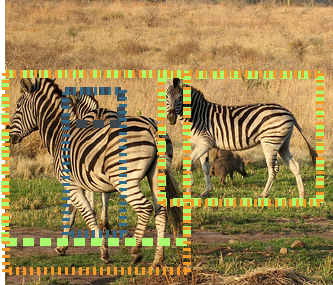}
        \caption{Problem Definition: Real label variation in LVIS \cite{lvis}. Different line style mean different raters.}
        \label{fig:real_variation}
    \end{subfigure}
    \hfill 
    \begin{subfigure}[t]{0.48\linewidth}
        \centering
        \includegraphics[width=\linewidth,trim={0 0 0 0},clip]{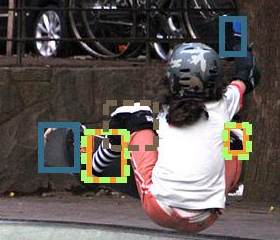}
        \caption{Validation Strategy: Synthetic label variations. Solid lines are synthetic annotations.}
        \label{fig:synthetic_variation}
    \end{subfigure}

    \vspace{0.6em} 

    \begin{subfigure}[t]{1.0\linewidth} 
        \centering
        \includegraphics[width=\linewidth,trim={0 0 0 0},clip]{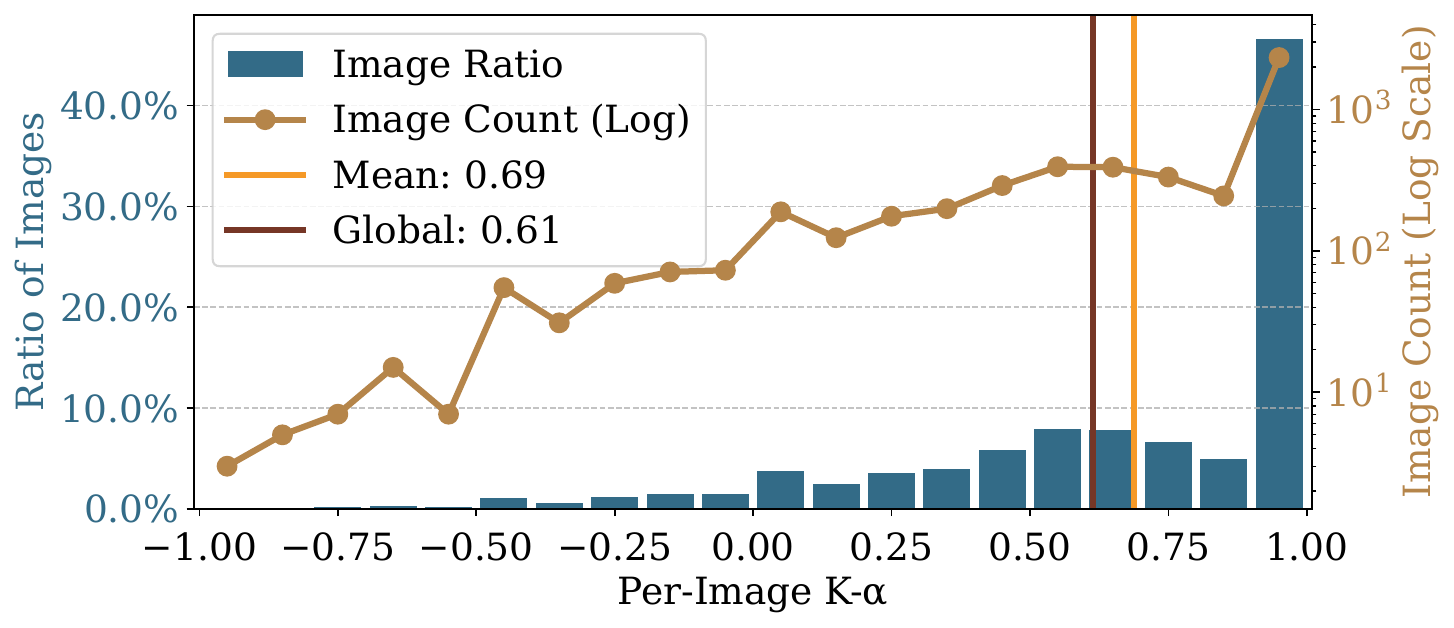}
        \caption{Method Application: Data quality measured for LVIS instance segmentation \cite{lvis} using \KalphaLOS{}, showing the general distribution of agreement across the entire LVIS consistency subset. Higher values indicate better agreement.}
        \label{fig:per_image_agreement_distribution}
    \end{subfigure}

    \vspace{0.6em} 

    \begin{subfigure}[t]{1.0\linewidth} 
        \centering
        \includegraphics[width=\linewidth,trim={0 0 0 0},clip]{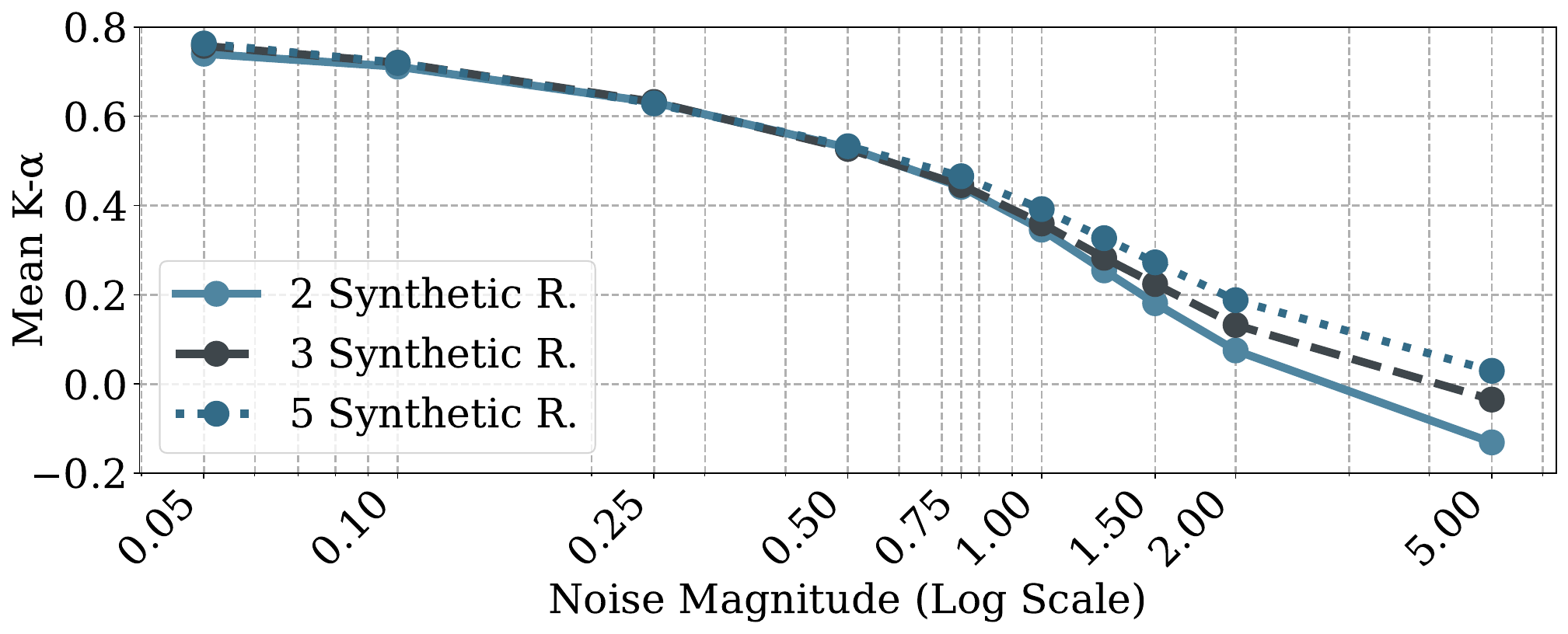}
        \caption{The Evidence: \KalphaLOS{} behavior when confronted with increasing magnitudes of synthetic label noise and a varying number of synthetic raters. As desired for a measurement of annotation consistency, it shows a monotonic decrease, full-range utilization, and no sudden jumps.}
        \label{fig:iaa_properties}
    \end{subfigure}

    \caption{Visual Abstract.}
    \label{fig:visual_abstract}
\end{figure}

%% file: sec/1_intro.tex
\section{Introduction}
\label{sec:intro}

Data quality fundamentally limits machine learning performance. Supervised learning remains the state-of-the-art approach for foundational tasks like object detection \cite{coco, lvis}. This approach relies on human-annotated data, which is considered the "gold standard" as a supervision signal. However, contradictions and ambiguity in this signal cap model performance during training \cite{agnew2023, mathet2012} and prevent algorithmic improvements, as evaluation on noisy labels becomes unreliable \cite{nassar2019, chachula2023}.

Tschirschwitz and Rodehorst \cite{tschirschwitz2025} quantify this issue using multi-rater datasets like LVIS-Consistency \cite{lvis}, TexBiG \cite{tschirschwitz2022}, and VinDr-CXR \cite{nguyen2022a}. By comparing multiple raters rather than assuming a single ground truth, they identify "label convergence" \cite{tschirschwitz2025} as an upper bound on model performance. Label convergence serves as a proxy for human-performance level, which effectively remains the measurable ceiling when evaluating models with data of similar quality.

Since current model performance \cite{zong2023, scylla-net} falls within the confidence interval of label convergence, evidence suggests that label quality, rather than architecture, acts as the fundamental bottleneck for progress in object detection. Stagnating benchmarks \cite{Codalab-COCO} on major computer vision datasets \cite{coco, lvis} further undermine claims of progress. To tackle this, researchers must judge the quality of labels. Yet, the vision community rarely assesses dataset quality. Even when assessed, the community relies on standard performance metrics such as mAP \cite{pfitzmann2022} or F1-score \cite{lvis}. This is detrimental. As Braylan \etal \cite{braylan2022} argue, these metrics fail to correct for chance agreement, hindering interpretability because the baseline for random agreement remains unknown.

Inter-Annotator Agreement (IAA) functions as a foundational component of the machine learning pipeline, from problem definition to evaluation. IAA does not assess the correctness of labels against an unknown truth, but rather the consistency between annotators, particularly for ill-defined tasks \cite{braylan2022}. IAA metrics are essential for refining guidelines and detecting systematic errors during dataset creation. They also enable downstream analyses, such as assessing class difficulty \cite{nassar2019} or annotator vitality \cite{nassar2019}, and post-creation evaluating label convergence \cite{tschirschwitz2025}. However, due to the wide variety of annotation tasks in Computer Vision (CV), no single IAA method has achieved adoption, leading to the inappropriate use of statistics and biased models \cite{braylan2022}.

To address this gap, we introduce our main contribution: \KalphaLOS{} (Krippendorff's $\alpha$ Localization Object Sensing), a novel meta-algorithm for assessing IAA in complex vision tasks. Unlike prior approaches that address specific aspects of agreement in isolation, \KalphaLOS{} unifies three critical capabilities into a single framework: (1) \textbf{Scope:} It extends easily to new tasks via a simple task-specific distribution analysis to identify a distance function; (2) \textbf{Depth:} It enables immediate downstream analysis without further adaptation; and (3) \textbf{Modularity:} Its components are interchangeable, allowing it to emulate previous methods as specific configurations \cite{amgad2022, tschirschwitz2022} or adapt to novel problem definitions.

Measuring the quality of an IAA method itself is challenging. An objective ground truth is absent by definition, making it impossible to verify the "correctness" of the instance correspondences that any IAA metric must establish. We solve this by introducing our second main contribution: a synthetic label noise generator. Built on complex, non-isotropic human variability derived from real-world data, this generator represents a significant leap over prior approaches that rely on axiomatic heuristics or black-box machine generation. This synthetic testbed provides necessary ground-truth control, allowing for statistically grounded experimental validation. We leverage this framework to empirically identify the optimal solver and cost function, establishing a robust standard configuration for \KalphaLOS{}.

%% file: sec/2_related_work.tex
\section{Related Work}
\label{sec:related_work}

\subsection{Foundations of Inter-Annotator Agreement}
While Cohen's \cite{cohen} and Fleiss' Kappa \cite{fleiss} laid the groundwork for quantifying IAA, the field has transitioned to the more flexible Krippendorff’s Alpha (\KripA) \cite{krippendorff}. This robust metric supports various data levels and arbitrary rater counts while natively handling missing data (implicit absence).

In machine learning, IAA adoption has diverged. The Natural Language Processing (NLP) community, which has long grappled with inherent linguistic ambiguity (\eg, in sentiment analysis \cite{davani2022} or Parts-Of-Speech (POS) tagging \cite{plank2014}), frequently employs IAA to validate datasets. Conversely, the CV community has historically prioritized dataset scale (\eg, ImageNet \cite{imagenet}, COCO \cite{coco}), often treating labels as a singular "gold standard". This practice overlooks rater variability, partly due to the high cost of acquiring repeated labels. A notable exception exists in medical imaging, where high stakes and expert disagreement are common. However, these works focus on semantic segmentation ("stuff") \cite{warfield2004,asman2011,ribeiro2019,yang2023a}, which does not address the distinct challenges of instance-based tasks ("things") that combine localization and classification.

\subsection{IAA for Object Detection}
\label{sec:related_work_iaa}

Adapting classic IAA to object detection is challenging because it requires resolving the instance correspondence problem, matching discrete "things" rather than pixels, before calculating consistency. Early attempts, such as Nassar \etal \cite{nassar2019}, applied \KripA's at pixel level. While this pioneered the evaluation of "rater vitality" in object detection, pixel-level evaluation treats distinct instances as continuous regions, or "stuff", failing to capture the discrete nature of detection tasks.

Recent approaches address this by decoupling the problem into two stages: first establishing spatial correspondence, then assessing classification agreement. Amgad \etal \cite{amgad2022} implement this for nuclei segmentation using Agglomerative Hierarchical Clustering (AHC), while Tschirschwitz \etal \cite{tschirschwitz2022} apply Sequential Hungarian Matching (SHM) on Document Layout Analysis. We adopt this "Localization First, Classification Second" principle not merely as a heuristic, but as a formal transformation step. By resolving spatial correspondence first, we transform a complex spatio-categorical problem into a standard nominal reliability matrix. This architecture allows \KripA{} to serve as the terminal metric. 
As the terminal metric remains a nominal $\alpha$, we can apply robust statistical tools (downstream analysis), such as rater vitality, class difficulty scoring, intra-annotator consistency, and collaboration clustering \cite{nassar2019, tschirschwitz2022}, to any CV task without task-specific re-implementation. While we primarily compare our method against object detection baselines in this work, this is solely because other complex tasks, such as pose estimation, currently lack established IAA implementations entirely.

Within this framework, choosing a valid localization distance function remains a challenge. Braylan \etal \cite{braylan2022} argue that a valid distance function must statistically distinguish between Observed Disagreement ($D_o$) and Expected Disagreement ($D_e$). However, because they employ complex, unified distance functions, they argue that standard metrics like \KripA{} become uninterpretable. Consequently, they propose abandoning \KripA{} in favor of ad-hoc distributional measures (\eg, $\sigma$ or the KS-statistic). We contend that discarding the mathematical foundation of \KripA, which natively handles small sample sizes, missing data, and arbitrary numbers of raters, is an over-correction. Instead, \KalphaLOS{} adopts Braylan’s distributional validation as a data-driven calibration mechanism to derive the optimal task-specific metric and threshold $\tau^*$. By deriving these components directly from the statistical separation between $D_o$ and $D_e$, we eliminate arbitrary hyper-parameter tuning in favor of principled, data-driven configuration. Pairing this validated configuration with the robust two-stage process allows us to retain the interpretability and statistical rigor of the established $\alpha$ coefficient.

\subsection{Datasets with Repeated Labels}

Our analysis relies on public datasets with repeated human annotations, including LVIS \cite{lvis}, TexBiG \cite{tschirschwitz2022}, and VinDr-CXR \cite{nguyen2022a}. We exclude datasets such as Open Images/COCO Reannotated \cite{ma2022} and NuCLS \cite{amgad2022} from our localization analysis, as their use of automated proposals confounds the study of pure human variability. See the Appendix (\cref{sec:dataset_selection}) for more details.

\begin{figure*}
  \centering
   \includegraphics[width=1.0\linewidth]{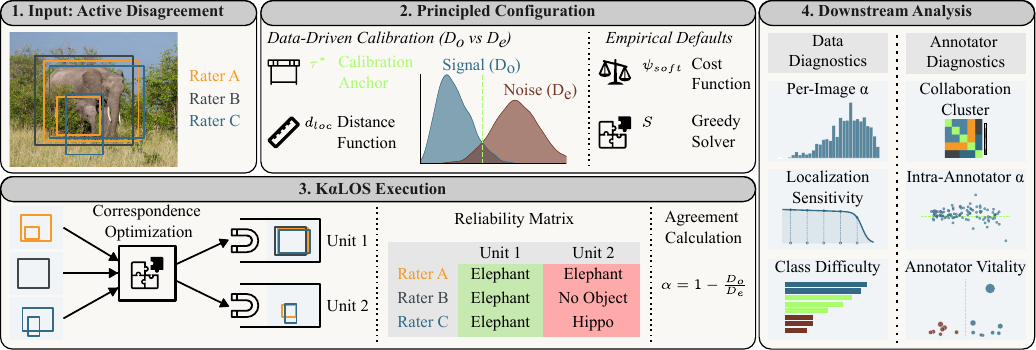}

   \caption{The \KalphaLOS{} Framework. (1) Input: The pipeline processes active disagreement between raters. (2) Principled Configuration: Data-driven calibration selects the optimal distance function ($d_{\text{loc}}$) and threshold ($\tau^*$) by maximizing the statistical separation between observed signal ($D_o$) and chance noise ($D_e$). (3) Execution: A solver transforms continuous spatial data into a discrete reliability matrix. Crucially, the completeness assumption encodes missed instances as \textsc{no\_object}, treating them as existence disagreement (false positives/negatives) rather than missing data. (4) Downstream Diagnostics: This standardization enables rigorous quality checks, visualizing Localization Sensitivity, Class Difficulty, and Annotator Vitality ($V_r$).
}
   \label{fig:kalos}
\end{figure*}

\subsection{Validation Gap: Synthetic Noise Generation}
\label{sec:related_work_synthetic_noise_generation}

Validating IAA metrics is uniquely challenging due to the absence of objective ground truth. Prior methods don't validate at all \cite{tschirschwitz2022}, rely on indirect proxies like downstream model performance \cite{nassar2019}, which introduce architectural confounds, or comparisons against expert aggregated labels \cite{amgad2022}, which contradict the inherent subjectivity of the task. While simulation-based frameworks \cite{braylan2022} allow for distributional analysis to rank distance functions such approaches often depend on axiomatic heuristics or black-box machine-generated labels that fail to capture the structured, non-isotropic nature of human error. For instance, existing noise models often simulate only localization offsets \cite{agnew2023, bar2023, liu2022a}, only classification errors \cite{liu2022}, or simple combinations \cite{li2020d, chadwick2019}. A critical flaw in many of these works, including recent multi-error injectors \cite{chachula2023, chan2023}, is the assumption of a uniform noise distribution, or the reliance on machine predictions that mimic algorithmic rather than human bias. For an extensive comparison see Appendix \cref{sec:prior_work}.

This creates a fundamental "Validation Gap": verifying an agreement score requires a truth that is absent by definition. Optimizing a metric to match a specific dataset's consensus risks overfitting to specific distributional biases rather than measuring agreement generally. To bridge this gap, we introduce a novel \textit{empirically derived} noise generator. Building on the philosophy of the Corpus Shuffling Tool (CST) \cite{mathet2012}, we model complex error distributions, including cardinal bias and semantic confusion, directly from real multi-rater datasets \cite{lvis,nguyen2022a,tschirschwitz2022}. This synthetic testbed provides the controlled ground truth necessary to verify \KalphaLOS{}'s properties.

%% file: sec/3_kalos.tex
\section{K\texorpdfstring{$\alpha$}{α}LOS: A Meta-Algorithm for IAA}
\label{sec:kalos_all}


\KalphaLOS{} is explained with a triple threat: (1) a visual overview in \cref{fig:kalos}, (2) textual descriptions and (3) formalized with equations in the following subsections.

\subsection{Problem Formulation and Desirable Properties}

Let an instance's unknown true label be $y_{ij} = (b_{ij}, c_{ij})$, representing its localization $b_{ij}$ (\eg, bounding box) and class $c_{ij}$ for instance $j$ in image $i$. We define the set of all true instances as $Y_i = \{y_{ij}\}_{j=1}^{N_i}$. In practice, we only observe approximations, $\Tilde{y}_{ik}^{r} = (\Tilde{b}_{ik}^{r}, \Tilde{c}_{ik}^{r})$ provided by a rater $r \in \{1,\dots,R\}$, defined as the set of annotations $\Tilde{Y}_i^r = \{\Tilde{y}_{ik}^{r}\}_{k=1}^{M_i^r}$ (\cref{fig:kalos}.1). When an approximation $\Tilde{y}_{ik}^{r}$ deviates significantly from its corresponding (and unknown) true label $y_{ij}$, the model receives an incorrect supervision signal.

Such deviations, or \textbf{disagreements}, manifest at several levels. The first is \textit{localization disagreement}, defined by a distance between spatial components:
\begin{equation}
    d_{\text{loc}}(b_{ij}, \Tilde{b}_{ik}^{r}) = 1 - \operatorname{IoU}(b_{ij}, \Tilde{b}_{ik}^{r}), 
\end{equation}
\begin{equation}
\label{eq:bbox_iou}
\operatorname{IoU}(b_{ij}, \Tilde{b}_{ik}^{r}) = 
\frac{|b_{ij} \cap \Tilde{b}_{ik}^{r}|}{|b_{ij} \cup \Tilde{b}_{ik}^{r}|}.
\end{equation}

The second, \textit{classification disagreement}, captures categorical differences:
\begin{equation}
d_{\text{cls}}(c_{ij}, \Tilde{c}_{ik}^{r}) =
\begin{cases}
0, & \text{if } c_{ij} = \Tilde{c}_{ik}^{r} \\
1, & \text{if } c_{ij} \neq \Tilde{c}_{ik}^{r}
\end{cases}
\end{equation}

Finally, \textit{unmatched instances} occur when an object is entirely missing (false negative, FN) or introduced incorrectly (false positive, FP). This occurs when an instance exists in one set but not the other:
\begin{equation}
\begin{split}
    &(y \in Y_i \text{ has no match in } \Tilde{Y}_i^r) \quad \text{or} \\
    &(\Tilde{y} \in \Tilde{Y}_i^r \text{ has no match in } Y_i)
\end{split}
\end{equation}

\noindent\textbf{Properties:}
To be effective for complex vision tasks, a modern IAA metric must satisfy several key properties to ensure it measures signal rather than noise:

\begin{itemize}
    \item \textbf{Monotonicity:} The score must monotonically decrease as the magnitude of label noise increases. This is the minimal requirement for validity.
    \item \textbf{Full-Range Utilization:} The metric should use its full scale (\eg, $[-1, 1]$) and not exhibit asymptotic behavior, which obscures meaningful differences at high or low agreement levels.
    \item \textbf{Sensitivity:} The metric must remain sensitive to changes in relevant areas, particularly subtle disagreements in high-agreement regimes. It should not "flatten out" prematurely, as this would mask important inconsistencies in dataset quality.
\end{itemize}

\subsection{Core Design Principles}
\label{sec:core_principles}

\KalphaLOS{} is founded on two core principles that transform the intractable continuous spatio-categorical problem into a solvable nominal form.

\noindent\textbf{1. Decoupling for Extensibility: Localization First.}
We adopt the established two-stage standard, solving localization first to establish correspondence before assessing agreement. This decoupling ensures generalizability, as adapting the framework to new tasks requires only defining a task-specific localization distance function. However, to resolve semantic ambiguities in spatially overlapping instances, such as one instance with two semantic similar classes (\cref{fig:cow_calf_figure}), we employ an extensible cost function $\psi_{soft}$ that incorporates class sensitivity information during the matching phase to disambiguate instances.

\noindent\textbf{2. Existence Disagreement \& The Completeness Assumption.}
We formally distinguish between \textit{active disagreement} and \textit{missing data}. Under the completeness assumption, we posit that raters find all instances they are assigned. Consequently, if a rater provides no annotation for a discovered unit, we classify this as an explicit \textsc{no\_object} decision (Active Disagreement), thereby penalizing FN and FP. Crucially, we reserve true missing data for unassigned tasks, such as scenarios where raters only annotate specific classes (\eg Open-World Detection). This distinction leverages the native ability of \KripA{} to handle missing values (implicit absence) while rigorously calculating reliability on the constructed matrix.

\subsection{The \KalphaLOS{} Pipeline}
\label{sec:kalos_pipeline}

\KalphaLOS{} transforms raw annotations in three steps into a reliability matrix.

\noindent\textbf{Step 1: Principled Configuration (\cref{fig:kalos}.2).}
To ensure the metric measures signal rather than noise, we split the configuration into fixed architectural components and adaptive, data-driven parameters.
\begin{itemize}
    \item \textbf{Empirically Validated Defaults (Signal Correspondence):} We fix the mechanical components to standardize the metric across tasks. Based on our synthetic validation (\cref{sec:kalos_exp_components}), we employ the Greedy solver ($\mathcal{S}$) and the semantically-sensitive cost function ($\psi_{\text{soft}}$). These components are not tuned per dataset. They are selected because they empirically maximize the recovery of annotator signal from complex, non-isotropic noise.
    
    \item \textbf{Data-Driven Calibration (Task Adaptation):} \KalphaLOS{} calibrates the localization function ($d_{\text{loc}}$) and the threshold ($\tau$) to the inherent statistics of the dataset. Both depend on a distributional analysis of \textit{Observed Disagreement} ($D_o$), representing disagreement between raters on the same image (signal), versus \textit{Expected Disagreement} ($D_e$), modeling chance by comparing raters across different images (noise). 
    Following the distributional validation logic of Braylan \etal \cite{braylan2022} (recap \cref{sec:braylan}), we select the function $d_{\text{loc}}$ (\eg, IoU) that maximizes the Kolmogorov-Smirnov (KS) separation between $D_o$ and $D_e$. We then derive the \textit{calibration anchor} $\tau^*$ via a Bayesian decision rule. Modeling $D_o$ and $D_e$ as probability densities $f$, $\tau^*$ identifies the crossover point where signal likelihood equals chance:
    \begin{equation}
        \label{eq:bayes_boundary}
        \tau^* = \{ \delta \in \mathbb{R} \mid f_{D_o}(\delta) = f_{D_e}(\delta) \}
    \end{equation}
    Rather than a hard limit, $\tau^*$ marks the aggregate transition from stochastic correspondence to stable consensus (see also calibration robustness analysis in \cref{sec:robustness_analysis}). Consequently, deviations from this anchor shift the evaluative scope: (a) configurations below $\tau^*$ measure fundamental existence agreement (detection), (b) whereas configurations above $\tau^*$ isolate geometric precision (localization).
\end{itemize}

\noindent\textbf{Step 2: Execution. (\cref{fig:kalos}.3)}
For each image $i$, \KalphaLOS{} processes the annotations to establish correspondence:
\begin{itemize}
    \item \textbf{Pairwise Cost Calculation:} \KalphaLOS{} computes the cost matrix using the semantically-sensitive cost function $\psi_{\text{soft}}$ (Eq. \ref{eq:category_lenient}). Crucially, any potential match where the distance $d > \tau$ is strictly discarded, effectively creating a sparse cost matrix to prevent implausible associations.
    \begin{equation}
        \psi_{\text{soft}}(d_{\text{loc}}, d_{\text{cls}}) =
        \begin{cases}
        -(1 - d_{\text{loc}}) - 1, & \text{if } d_{\text{cls}} = 0 \\
        -(1 - d_{\text{loc}}), & \text{if } d_{\text{cls}} = 1
        \end{cases}
        \label{eq:category_lenient}
    \end{equation}
    
     \item \textbf{Correspondence Optimization:} The solver $\mathcal{S}$ receives the pairwise costs $C_{kl}$ for matching annotations between raters. It finds the correspondence set $M^*$ that minimizes the total cost subject to cycle-consistency (\ie, if $k \leftrightarrow l$ and $l \leftrightarrow m$, then $k \leftrightarrow m$):
    \begin{equation}
        M^* = \underset{M}{\operatorname{argmin}} \sum_{(k, l) \in M} C_{kl}
    \end{equation}
    This process groups individual annotations into a set of disjoint ``units'' $U_i$, where each unit represents a single, agreed-upon instance.

    \item \textbf{Reliability Matrix Construction:} We transform the disjoint units $U_i$ into a per image matrix where rows represent raters and columns represent discovered units. Empty cells are filled based on the Completeness Assumption: if a rater $r$ was assigned to the image but did not annotate the unit, the cell is marked as \textsc{no\_object}. If the rater was not assigned, it is marked as \textsc{nan}.
\end{itemize}

\noindent\textbf{Step 3: Terminal Agreement.}
We compute the final score using \KripA{} on the constructed reliability matrix for each image (recap \cref{sec:kalpha}). To account for any sample size $n$ and missing data, we calculate $\alpha$ via the coincidence matrix. Let $o_{cc}$ be the count of observed pairable values where both raters agreed on class $c$. We define the total counts per category as (frequency) $n_c$ and the total pairable observations as $n$.
The final score effectively normalizes the observed disagreement ($D_o$) against the expected disagreement ($D_e$):
\begin{equation}
    \alpha = 1 - \frac{D_o}{D_e} = \frac{(n-1)\sum_c o_{cc} - \sum_c n_c(n_c-1)}{n(n-1) - \sum_c n_c(n_c-1)}
\end{equation}
The dataset wide primary score is the per image mean of $\alpha$'s. It is also possible to calculate a global $\alpha$ by concatenating the reliability matrix as a secondary score. A comparison and discussion is found in \cref{sec:global_vs_mean_alpha}.

\textbf{Interpretation:} The resulting score ranges from $1$ (perfect agreement) to $-1$ (systematic disagreement), with $0$ indicating random chance. Following standard conventions~\cite{amgad2022,landis1977a}: $\alpha \ge 0.8$ indicates near-perfect agreement, $\alpha \ge 0.6$ is substantial, $\alpha \ge 0.4$ is moderate, and $\alpha < 0$ implies systematic errors in the guideline or annotation process.

Example notation for a finished \KalphaLOS{} configuration:
\begin{equation}
\mathrm{K}\alpha\mathrm{LOS}_{\big(d_{\mathrm{loc}},\,\tau{=}0.5,\,\mathcal{S}{=}\text{Greedy},\,\psi_{\text{soft}}\big)}
\end{equation}

\textbf{Downstream Tasks:} A primary advantage of the \KalphaLOS{} Meta-Algorithm is its ability to standardize complex visual tasks into a nominal reliability matrix. This transformation enables the application of a diagnostics toolkit (\cref{fig:kalos}.4) that are otherwise incompatible with raw spatial data. By re-computing $\alpha$ on filtered or permuted subsets of this matrix, the framework facilitates granular diagnostics that extend beyond a single summary score. Detailed discussions of these downstream applications are provided in \cref{sec:downstream_tasks}.

\textbf{Examples:} We demonstrate the versatility of \KalphaLOS{} across various domains in \cref{sec:application_of_kalos}. These applications include extensions to tasks such as instance segmentation (\cref{sec:instance_segmentation}), 3D volumetric instance segmentation (\cref{sec:3D}), and pose estimation (\cref{sec:pose_estimation}), practical utilization of data-driven calibration, as well as a demonstration of downstream analysis.

%% file: sec/4_experimental_validation_framework.tex
\section{Experimental Validation}
\label{sec:framework}

\textbf{Goal and Scope.} This framework verifies that the \KalphaLOS{} pipeline accurately measures agreement and satisfies the properties defined in \cref{sec:kalos_all} (monotonicity, full-range utilization, and sensitivity). To ensure robust application, we validate the pipeline configuration, specifically the correspondence solver and cost function, that best identifies the intended annotator signal.

\textbf{The Validation Gap.} Validating an IAA metric is fundamentally circular: verification requires an objective ground truth that is absent by definition in the tasks that IAA is designed to measure. Prior validation methods fail to address this (\cref{sec:related_work_synthetic_noise_generation}), as we cannot objectively test if a metric correctly penalizes human-like disagreement without a reference that mimics human variability while retaining known parameters \cite{braylan2022}.

\textbf{Solution: An Empirically-Derived Testbed.} We bridge this gap by developing a controlled noise generator that synthesizes "ground truth" disagreement. This testbed offers a superior validation environment: it combines the distributional realism of human annotators with the absolute control of synthetic generation.

\begin{figure*}
    \centering
    \includegraphics[width=1.0\linewidth]{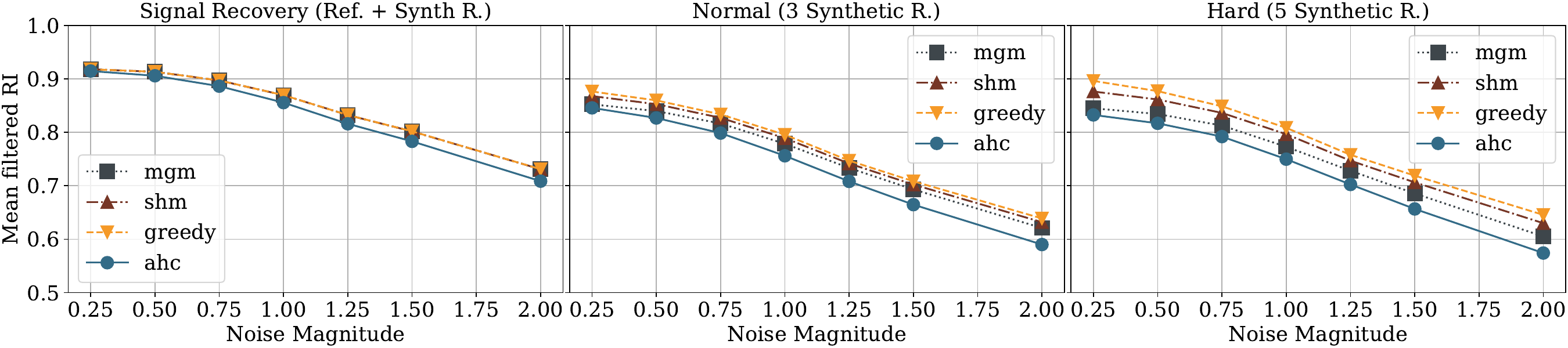}
    \caption{Performance of different correspondence solvers averaged across LVIS, TexBiG and VinDr-CXR. A simple filtered Rand Index variation was used due to the constrained clustering setup, where limited and partially overlapping clusters, together with non-exhaustive ground-truth coverage, rendered more advanced evaluation metrics infeasible or unreliable for assessing correspondence accuracy.}
    \label{fig:ri_lenient_all}
\end{figure*}

\begin{itemize}
    \item  \textbf{Empirical Noise Modeling} (details in \cref{sec:noise_generator}). We reject the common assumption of uniform error \cite{chachula2023}. Instead, we propose a methodological leap in generator design that escapes the use of axiomatic heuristics or "black-box" machine generation: a strict \textit{Empirical Data → Model Fit → Statistical Test} loop. Through an iterative process of hypothesis testing on multi-rater datasets \cite{nguyen2022a, lvis, tschirschwitz2022}, we identified the specific distributions that govern human error. This yielded a Dual-Layer Framework: (1) Parametric Drivers that capture the heavy-tailed, size-dependent nature of geometric variability and error rates; and (2) Validated Proxies that utilize foundation models to sample semantically and visually ambiguous "ghost" objects, mimicking the non-random nature of false positives and category confusion. Finally, we compose these models into a coherent generator that adheres to a strict synthesis hierarchy (Unmatched → Topology → Category → Localization), ensuring that complex interactions between errors (\eg, signal saturation) are modeled realistically.
    
    \item \textbf{Component Evaluation.} Using this testbed, we isolate and evaluate the core components of \KalphaLOS{} to validate which components are best able to capture the intended annotation signal. We compare four correspondence solvers (Greedy, SHM \cite{tschirschwitz2022}, MGM \cite{kahl2025}, and AHC \cite{amgad2022}) and two cost functions ($\psi_{soft}$ and $\psi_{neg}$). Crucially, we evaluate previous methods as specific configurations of \KalphaLOS{}: SHM with $\psi_{neg}$ corresponds to \cite{tschirschwitz2022} and AHC with $\psi_{neg}$ corresponds to \cite{amgad2022}.
    
    \item \textbf{Property Verification.} We confirm that \KalphaLOS{} exhibits strict monotonicity, sensitivity to subtle disagreements and full range utilization, ensuring the metric behaves as intended on the synthetic testbed (\cref{sec:kalos_exp_components}) and on real datasets (\cref{fig:iaa_properties}, \cref{sec:application_of_kalos}).
\end{itemize}

%% file: sec/5_experiment_and_results.tex
\section{Experiments and Results}
\label{experiments_and_results}

Using the controlled synthetic noise environment from \cref{sec:framework}, we systematically assess \KalphaLOS{}  and its components across increasing noise levels and numbers of annotators to analyze correspondence accuracy, stability, and overall robustness.


\subsection{\KalphaLOS{} Components}
\label{sec:kalos_exp_components}

\textbf{Experiment 1 – High-Level Clustering Performance.} We evaluate which correspondence solver most accurately recovers inter-annotator cluster structures under increasing annotation noise. As shown in \cref{fig:ri_lenient_all}, all solvers perform near identical in the signal recovery (reference and one synthetic rater). However, when realistic noise and limited rater evidence are introduced (3 synthetic raters), performance drops sharply for all methods due to cross-cluster ambiguities. Increasing the number of raters (5 synthetic raters) partially restores performance through majority consensus. Across all conditions, Greedy and sequential Hungarian solvers yield the highest filtered Rand Index, indicating that locally optimal strategies better tolerate structured human-like noise, whereas state-of-the-art globally optimal MGM \cite{kahl2025} is overly conservative, favoring precision over recall. Agglomerative clustering performs worst across all noise levels due to uncontrolled merge propagation.

\begin{figure}
  \centering
   \includegraphics[width=1.0\linewidth]{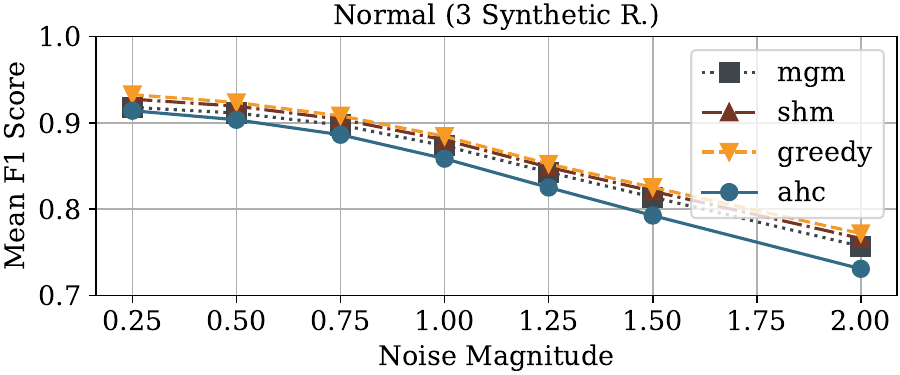}

   \caption{F1-score comparison. Greedy performs best.}
   \label{fig:f1_score}
\end{figure}

\textbf{Experiment 2 – Precision vs. Recall Analysis.} To understand whether solvers predominantly make merge errors (low precision) or split errors (low recall), we analyzed their precision–recall behavior. The precision curves are omitted, as all algorithms and cost-function combinations achieve uniformly high precision ($\approx$ 0.97–1.0), offering little visual discrimination. This indicates that none of the solvers produce substantial spurious correspondences; the dominant failure mode is missed matches. Consequently, meaningful differences arise in recall, where the more aggressive solvers (Greedy, SHM) consistently outperform the more conservative MGM, resulting in higher F1-scores \cref{fig:f1_score}. Based on these findings, we exclude AHC from further experiments and focus on Greedy, SHM, and MGM, using \cref{eq:category_lenient}, $\psi_{\text{soft}}$, as out cost function, which shows the best and most consistent RI and F1 performance (details in \cref{sec:cost_functions}).

\begin{figure*}
  \centering
   \includegraphics[width=1.0\linewidth]{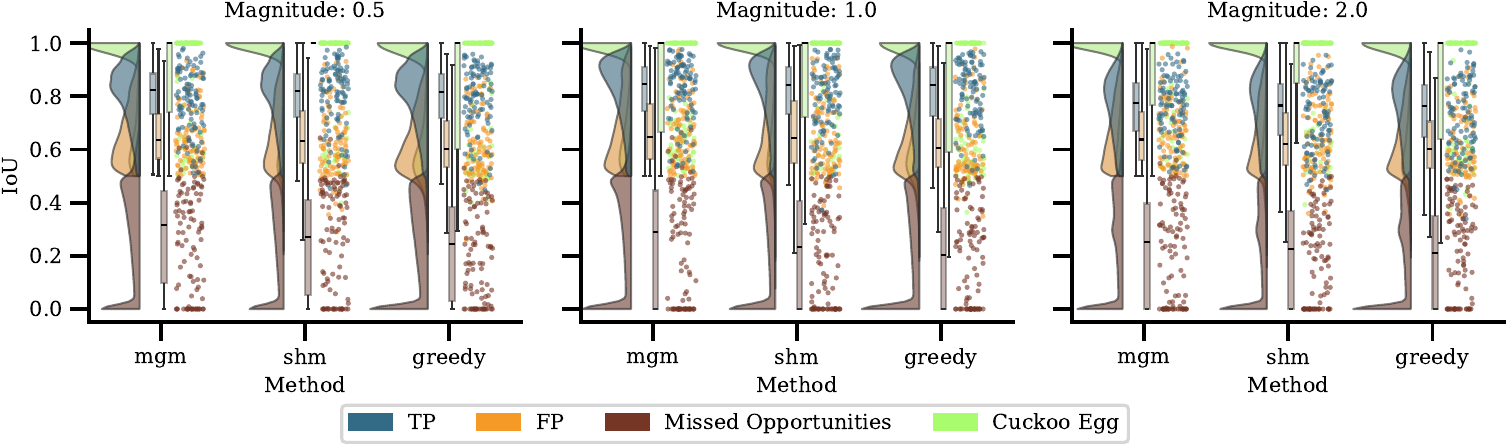}

   \caption{Failure-type comparison across correspondence solvers and noise magnitudes using raincloud plots. TP shows correctly matched pairs, Missed Opportunities are correct pairs the solver failed to match, and FP captures spurious matches where no true pair exists. Cuckoo Eggs indicate cases where a correct match was displaced by an incorrect one, effectively combining a FN with a FP and revealing when class disagreement overrides $\psi_{\text{soft}}$ \cref{eq:category_lenient} function.}
   \label{fig:failure_modes}
\end{figure*}

\textbf{Experiment 3 – Failure Case Analysis.} We analyze solver behavior using raincloud plots in \cref{fig:failure_modes} that decompose matches into true positives (TP), FP, missed opportunities, and “cuckoo eggs.” Greedy exhibits the broadest TP IoU distribution, indicating the strongest reach and willingness to accept difficult but correct matches, while MGM concentrates almost exclusively on high-IoU pairs, revealing a conservative, “cherry-picking” strategy. Missed opportunities complement this pattern: Greedy’s misses are mostly low-IoU pairs, while MGM and SHM still miss a noticeable fraction of high-IoU candidates. FP distributions remain small overall; Greedy shows few high-IoU FPs, indicating it is rarely fooled by plausible distractors inside the 0.5 threshold. Cuckoo eggs, where a correct match is displaced by an incorrect one, occur only rarely and mostly at high IoU, consistent with the intended behavior of the lenient cost function under class disagreement. Overall, Greedy captures a wider profile of valid correspondences with fewer severe errors.

\textbf{Experiment 4 – Stability to Permutations.} Finally, we test solver determinism by permuting raters and annotations. Greedy is the only method that remains perfectly stable on the most challenging real dataset (NuCLS \cite{amgad2022}), while MGM and SHM exhibit measurable instability (\eg, NuCLS ARI means: greedy 0.99998, mgm 0.9606, shm 0.9327). Combined with its superior recall and failure-case robustness, this establishes Greedy as the  recommended default solver in combination with \cref{eq:category_lenient} as the cost function. The principal selection for the distance function and valid range for $\tau$ remain task dependent to maintain extensibility to new tasks.

%
%

\subsection{\KalphaLOS{} Experiments}

\begin{figure}
  \centering
   \includegraphics[width=1.0\linewidth]{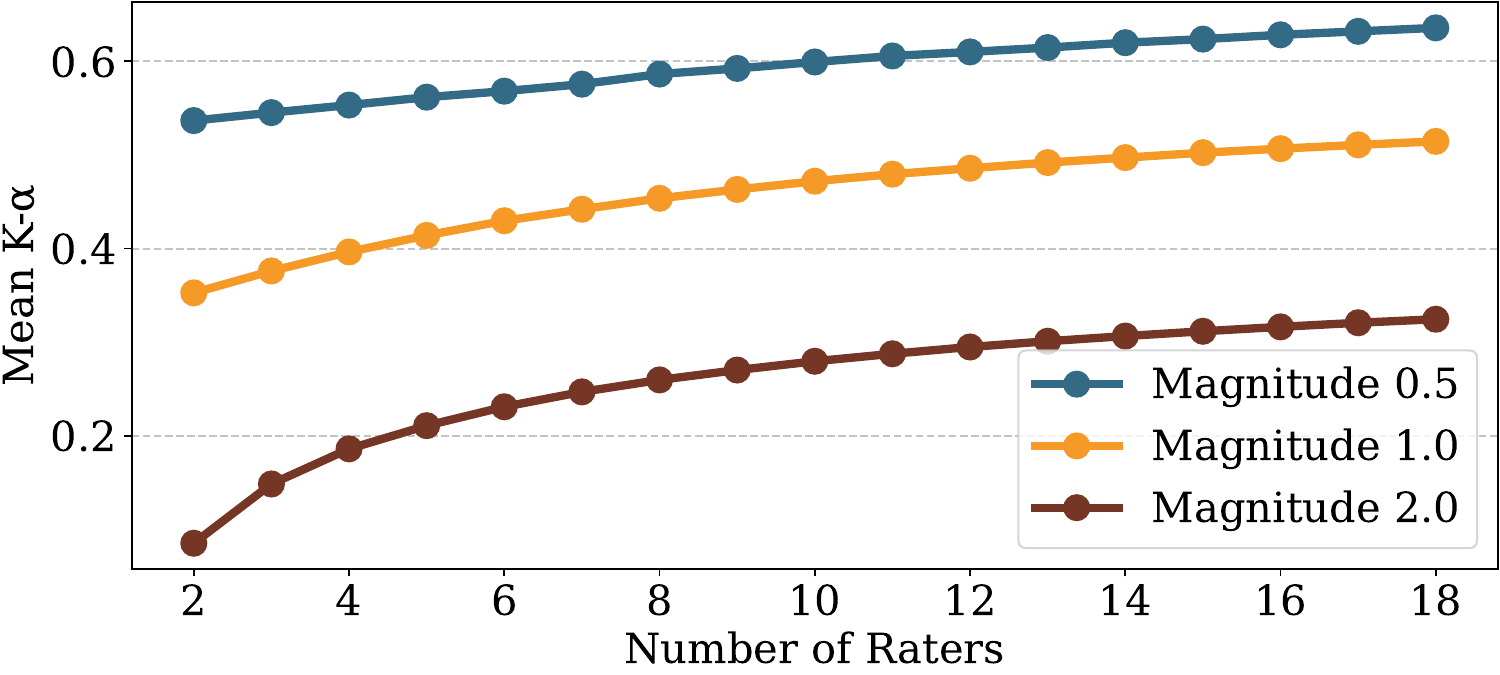}

   \caption{Return on investment for using more annotators. Majority voting principal applies. Annotators that generally follow the same annotation convention increase consensus, independent of increasing magnitude.}
   \label{fig:roi_iaa}
\end{figure}

\begin{figure}
  \centering
   \includegraphics[width=1.0\linewidth]{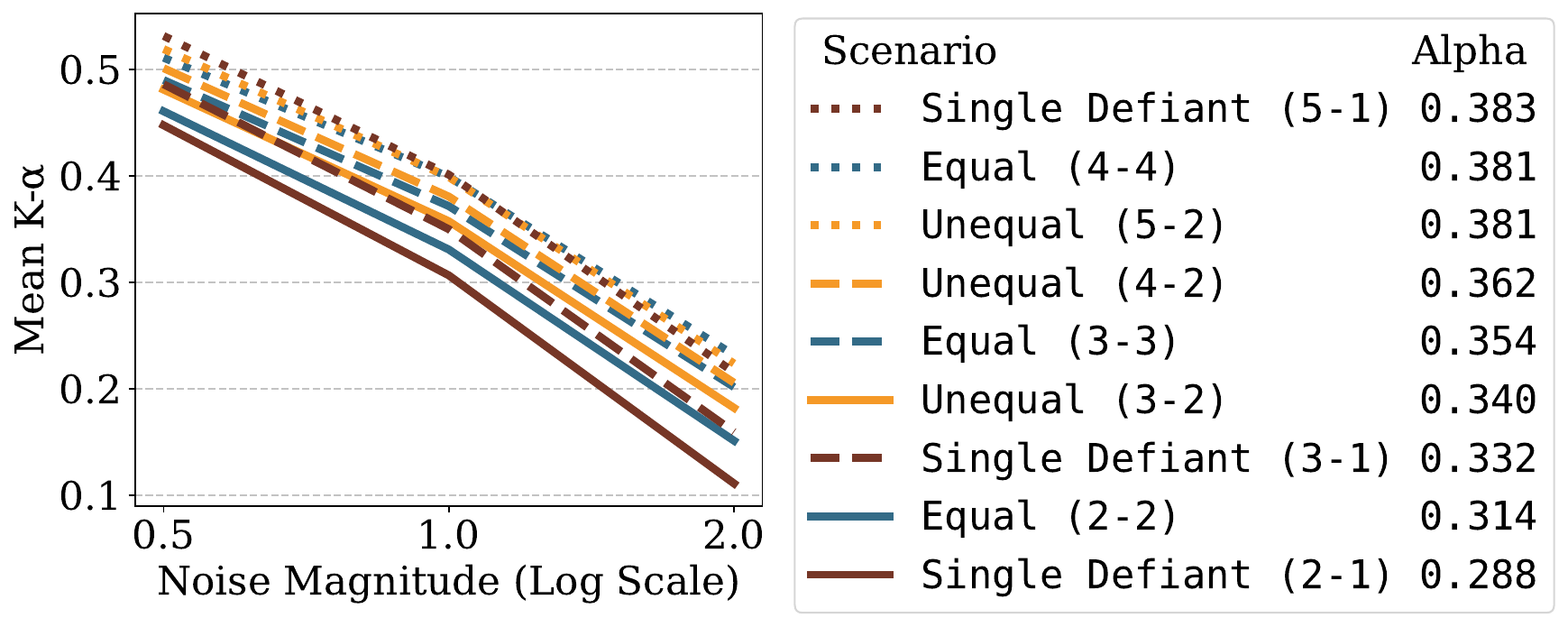}

   \caption{Collaboration-cluster size impact analysis between two groups of A and B raters, denoted as (A–B) for the group sizes. LVIS is used to generate synthetic annotations from two anonymous, interchangeable reference raters.}
   \label{fig:cc_iaa}
\end{figure}

The core properties of \KalphaLOS{} (\cref{sec:core_principles}) are highlighted in \cref{fig:iaa_properties} and are again validated in the following two experiments.

\textbf{Experiment 5 – Return on Investment for Additional Raters.} Using our empirical noise generator, we vary the number of raters from 2 to 18 at fixed noise magnitudes (\cref{fig:roi_iaa}). Mean \KalphaLOS{} scores increase monotonically with more raters, with steep gains for the first few raters and clear diminishing returns beyond roughly 6–8 raters. This mirrors the redundancy analysis of NuCLS \cite{amgad2022}, where AUROC (a metric used in NuCLS) for inferred crowd labels saturates after $\approx$ 6 non-pathologists per field of view, while still benefiting from majority voting among raters who follow similar annotation conventions.

\textbf{Experiment 6 – Impact of Collaboration Clusters.} We next vary the composition of “schools of thought” while controlling the total number of raters (\cref{fig:cc_iaa}). Scenarios with a large, consistent majority and a single defiant annotator (\eg, 5–1) achieve the highest agreement, as the majority effectively outvotes the outlier. Configurations with equally sized or sizable opposing clusters (\eg, 3–3, 3–2, 2–2) show noticeably reduced \KalphaLOS{} scores, even when more raters are present. Thus, \KalphaLOS{} captures both the benefit of additional raters and the degradation in consensus when multiple annotation conventions coexist.

For further downstream analysis, default configuration and applications of \KalphaLOS{} to different datasets and \textbf{task extensions}, we refer to \cref{sec:application_of_kalos}. Code for \KalphaLOS{} will be made available for research and commercial use.

%% file: sec/6_conclusion.tex
\section{Conclusion}
We introduced \KalphaLOS{}, a meta-algorithm that unifies the evaluation of inter-annotator consistency across instance-based vision tasks. By determining a valid distance function via distributional analysis, the framework adapts to diverse domains beyond object detection. Applied to subsets of repeated labels, \KalphaLOS{} serves as a practical diagnostic tool: it identifies ambiguous guidelines, isolates problematic raters, and, when combined with label-convergence analysis \cite{tschirschwitz2025} establishes realistic performance ceilings for downstream models.

However, the framework operates under three constraints. First, the decoupled architecture relies on statistical separability between localization and classification, requiring narrow threshold calibration in spatially constrained tasks like laboratory pose estimation. Second, empirical validation is currently limited to object detection due to data availability; conclusions regarding solver stability in pose or volumetric domains remain extrapolations. Finally, the method strictly requires multi-rater metadata, preventing post-hoc quality assurance on legacy single-annotator datasets without additional sampling.

To validate this framework, we developed an empirically grounded noise generator that models human error distributions rather than heuristic perturbations. While this testbed confirms the metric's monotonicity and stability, the current generation process lacks coupling to image content. Future work must incorporate visual uncertainty maps to enable the broader use of these synthetic labels for model training and augmentation.

\section*{Acknowledgments}
Funded by the Deutsche Forschungsgemeinschaft (DFG, German Research Foundation) – 556314568.


%% file: sec/7_appendix.tex
\clearpage
\setcounter{page}{1}
\maketitlesupplementary

\input{sec/foundations}

\input{sec/datasets}

\input{sec/noise_generator}

\input{sec/cost_functions}

\input{sec/robustness_analysis}

\input{sec/global_vs_mean}

\input{sec/downstream_tasks}

\input{sec/kalos_applications}

%% file: sec/foundations.tex
\section{Foundations}
\label{sec:foundations}

\subsection{Recap of Krippendorff's Alpha}
\label{sec:kalpha}

\textbf{Goal.} To quantify inter-annotator agreement using a metric that is robust to missing data, variable numbers of raters, and small sample sizes. Unlike simple percentage agreement, \KripA{} corrects for chance, ensuring that the reported score reflects genuine consensus rather than random coincidence.

\textbf{Problem.} Standard correlation metrics require fully paired data where every rater annotates every item. In complex vision tasks, this assumption rarely holds. We require a method that distinguishes between a rater who was \textit{not assigned} a task (implicit absence) and a rater who \textit{saw} the image but found nothing (explicit absence/active disagreement).

\textbf{Solution.} We utilize the nominal form of \KripA{}. The calculation proceeds in two steps: transforming the reliability matrix into a coincidence matrix, and computing the ratio of observed to expected disagreement.

\noindent\textbf{1. The Coincidence Matrix.}
The pipeline first flattens the reliability matrix (columns=units $u$, rows=raters $r$) into a coincidence matrix. For every unit $u$, we generate all possible unique pairs of labels provided by the assigned raters. The handling of empty cells in the reliability matrix is the critical differentiator:
\begin{itemize}
    \item \textbf{Implicit Absence (Missing Data):} If a rater provides no label because they were never asked to annotate that specific image or class, the entry is marked \texttt{NAN}. These entries generate no pairs and do not influence the score. This natively handles:
    \begin{itemize}
        \item \textit{Sparse assignment:} Images annotated by different subsets of raters.
        \item \textit{Open World / Class Subsets:} Scenarios where specific raters are assigned only specific class subsets (e.g., rater A labels ``animals'', rater B labels ``vehicles''). If Rater A does not annotate a ``car'', it is missing data, not disagreement.
    \end{itemize}
    \item \textbf{Explicit Absence (Active Disagreement):} If a rater is assigned to the image/class but provides no annotation for a discovered unit $u$, this is treated as the class \texttt{NO\_OBJECT}. A disagreement between \texttt{Class A} and \texttt{NO\_OBJECT} is treated mathematically identical to a disagreement between \texttt{Class A} and \texttt{Class B}.
\end{itemize}

\noindent\textbf{2. The Metric (\KripA{}).}
We compute the agreement $\alpha$ by comparing the observed coincidence of labels against the distribution expected by chance.

\begin{equation}
\label{eq:krippendorff_recap}
\alpha = 1 - \frac{D_o}{D_e} = \frac{(n-1) \sum_{c} o_{cc} - \sum_{c} n_c(n_c-1)}{n(n-1) - \sum_{c} n_c(n_c-1)}
\end{equation}

\noindent\textbf{Where:}
\begin{itemize}
    \item $n$: The total number of values paired across all units in the coincidence matrix.
    \item $n_c$: The total count of class $c$ (frequency) across the entire dataset.
    \item $o_{cc}$: The observed count of pairs where both raters agreed on class $c$ (the diagonal of the coincidence matrix).
    \item $D_o$: Observed disagreement (frequency of mismatched pairs).
    \item $D_e$: Expected disagreement (frequency of mismatches if raters assigned labels randomly based on the population distribution $n_c$).
\end{itemize}

\textbf{Intuition.} The metric operates as a ratio of signal to noise. If raters agree perfectly, the observed matches ($o_{cc}$) maximize, resulting in $\alpha = 1$. If raters agree only as often as a randomized shuffle of their collective labels would predict, $D_o \approx D_e$, resulting in $\alpha = 0$. Systemic disagreement (worse than chance) yields $\alpha < 0$.

\subsection{Recap: Distributional Validation}
\label{sec:braylan}

 In complex annotation tasks, choosing a distance function $d$ (e.g., IoU, L2, OKS) is often heuristic. Braylan \etal \cite{braylan2022} propose eliminating this arbitrariness by validating $d$ against the dataset’s inherent statistical properties. They posit that a valid distance function must maximize the separation between two distributions:
 \begin{enumerate}
     \item Observed Disagreement ($D_o$): The pairwise distances between different raters annotating the same item (representing signal).
     \item Expected Disagreement ($D_e$): The pairwise distances between raters annotating different items (representing chance/noise).
 \end{enumerate}

If $d$ is valid, $D_o$ should be stochastically smaller than $D_e$. This separation is quantified using the Kolmogorov-Smirnov (KS) statistic: \begin{equation} 
    KS = \sup_x |F_{D_o}(x) - F_{D_e}(x)| 
\end{equation} 
where $F$ is the empirical cumulative distribution function. In \KalphaLOS{}, we adopt this logic not to replace $\alpha$, but as a calibration step (\cref{sec:kalos_pipeline}) to empirically select the task-optimal $d_{loc}$ and the transition threshold $\tau^*$ where signal becomes distinguishable from noise.

\input{sec/dataset_table}

\subsection{Notations}
\label{sec:notations}

\textbf{Ground Truth}
\begin{itemize}
    \item $y_{ij} = (b_{ij}, c_{ij})$: The unknown true label for instance $j$ in image $i$.
    \item $Y_i = {y_{ij}}_{j=1}^{N_i}$: The set of all true instances in image $i$.
\end{itemize}
\textbf{Annotations (Observed Data)}
\begin{itemize}
    \item $\Tilde{y}_{ik}^{r} = (\Tilde{b}_{ik}^{r}, \Tilde{c}_{ik}^{r})$: An annotation from rater $r$.
    \item $\Tilde{Y}_i^r$: The set of all annotations from rater $r$ for image $i$.
    \item $r$: A specific rater.
    \item $R$: The set of all raters in the dataset.
    \item $R_i$: The subset of raters assigned to image $i$.
\end{itemize}
\textbf{Disagreement, Distance and Cost}
\begin{itemize}
    \item $d_{\text{loc}}$: Localization distance (\eg, $1 - \text{IoU}$).
    \item $d_{\text{cls}}$: Classification distance (0 for match, 1 for mismatch).
    \item $d$: A general distance function (used in \KalphaLOS{} configuration).
    \item $\psi$: A general cost function for the correspondence solver.
    \item $\psi_{\text{soft}}$: A specific cost function that incorporates class information.
    \item $C_{kl}$: The cost of matching annotation $k$ with annotation $l$.
\end{itemize}
\textbf{\KalphaLOS{} Pipeline Components:}
\begin{itemize}
    \item $\alpha$ or \KripA{}: Krippendorff's Alpha.
    \item $\tau$: Localization threshold.
    \item $\mathcal{S}$: The correspondence solver (\eg, Greedy).
    \item $M^*$: The optimal correspondence set found by the solver.
    \item $U_i$: The set of disjoint "units" (clusters) for image $i$.
    \item $\text{\textsc{no\_object}}$: Special category for a rater's absence of an annotation in a unit. Active disagreement.
\end{itemize}
\textbf{Noise Generator:}
\begin{itemize}
    \item $\lambda$: Magnitude of the noise generator.
\end{itemize}

%% file: sec/dataset_table.tex
\begin{table*}[t]
\caption{Overview of datasets with repeated annotations. Bold names are the ones used for our main analysis.}
\label{tab:datasets}
\centering
\renewcommand{\arraystretch}{1.25}
\small

\rowcolors{2}{gray!10}{white}

\begin{tabularx}{\textwidth}{@{}p{2.5cm}XXXXXX@{}}
\toprule
\textbf{Dataset} &
\textbf{Task(s)} &
\textbf{Amount of Raters per Image/Scene} &
\textbf{Annotator Type} &
\textbf{Rater Identification + Intra-Rater Evaluation} &
\textbf{Guideline and Annotation Process} &
\textbf{Automated Proposals} \\
\midrule

\textbf{TexBiG \cite{tschirschwitz2022}} &
Instance Segmentation &
3--4 (train), 5 (test) &
Mixed (Experts + Non-Experts) &
Yes + No &
Same &
no \\

\textbf{VinDr-CXR \cite{nguyen2022a}} &
Object Detection &
3 (train), 5 (test) &
Experts (Radiologists) &
Yes + No &
Same &
no \\

\textbf{LVIS Consistency Subset \cite{lvis}} &
Instance Segmentation &
2 &
Non-Experts &
No + No &
Same &
no \\

COCO Reannotated \cite{ma2022}&
Object Detection &
2 &
Non-Experts &
No + No &
Different &
yes \\

Open Images Reannotated \cite{ma2022} &
Object Detection &
2 &
Non-Experts &
No + No &
Different &
yes \\

NuCLS \cite{amgad2022}&
Instance Segmentation &
11--39 &
Mixed (Experts + Non-Experts) &
Yes + Yes &
Same &
yes \\

LIDC-IDRI (TCIA) \cite{armato2015lidc}&
Voxel Grid / 3D Volume &
4 &
Experts (Radiologists) &
No + No &
Different (see text) &
no \\

MARS \cite{segalin2021}&
Video Pose Estimation and Action Detection (behavior) &
5 (pose), 8 (behavior; no localization) &
Non-Experts (pose) + Experts (behavior) &
Yes + Yes &
Same &
no \\

\bottomrule
\end{tabularx}
\end{table*}

%% file: sec/datasets.tex
\section{Dataset Selection}
\label{sec:dataset_selection}

The datasets considered in this work had to satisfy the following criteria:
\begin{itemize}
    \item The task combines localization and classification.
    \item Annotations are fully human-created.
    \item The data are publicly accessible.
    \item A consistent annotation guideline is used across all annotations.
\end{itemize}

The core datasets that are eligible for at least parts of our study are summarized in \cref{tab:datasets}. They are used either (1) for synthetic data creation (usable for empirical noise analysis), (2) as reference annotations for synthetic data generation and thus \KalphaLOS{} validation, or (3) as examples for applications of \KalphaLOS{} (see \cref{sec:application_of_kalos}).

\begin{figure*}
  \centering
   \includegraphics[width=1.0\linewidth]{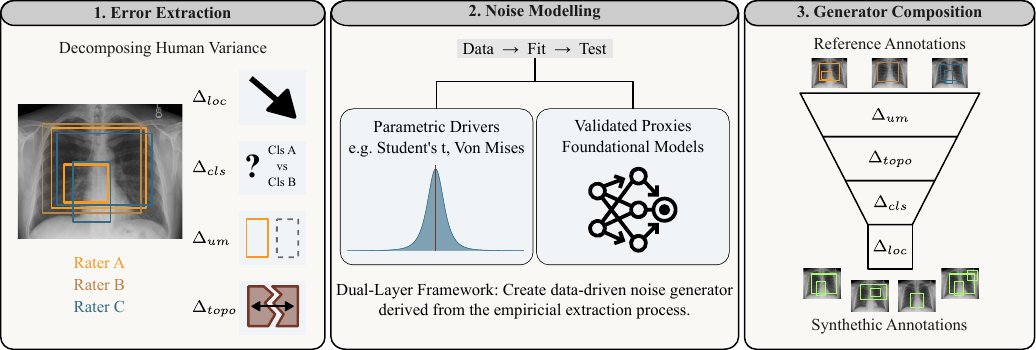}

   \caption{Three step noise generator: 1) Empirical data is extracted. 2) Data is used to fit parametric models or estimate them using foundational models as validated proxies. 3) The four separate components use a hierarchical composition to generate synthetic annotations based on reference annotations (\eg human annotated data).}
   \label{fig:noise_generator}
\end{figure*}

For completeness, we briefly discuss additional datasets that were considered but ultimately not used:
\begin{itemize}
    \item \textbf{KeypointNet} \cite{keypointnet}: designed for 3D keypoint detection, but only aggregated annotations are available, which makes it unsuitable for our multi-annotator analysis.
    \item \textbf{CorresPondenceNet} \cite{correspondencenet}: a 3D correspondence dataset where annotators mark sets of semantically consistent points across ShapeNet objects, providing human correspondence consensus rather than explicit part or keypoint labels, which makes it unsuitable for our instance-based IAA setting.
    \item \textbf{IBIS} \cite{ibis}: contains dental radiography landmarks, but the underlying data were no longer accessible at the time of our study.
    \item \textbf{MultiOrg} \cite{multiorg}: at the time of writing, the associated challenge was still ongoing, and the test images with repeated annotations were not yet publicly released.
\end{itemize}

%% file: sec/noise_generator.tex
\section{Empirical Noise Generator}
\label{sec:noise_generator}

\textbf{Goal and Scope.} Validating agreement metrics requires an objective ground truth that mimics the complexity of human error. As Agnew \etal \cite{agnew2023} note, the true distribution of annotation uncertainty remains unknown. Consequently, most existing methodologies rely on axiomatic heuristics or black-box machine generation, which fail to capture the heavy-tailed, non-isotropic nature of human disagreement.
\textbf{The Dual-Layer Framework.} We address this gap by constructing a data-driven noise generator derived from the empirical extraction process of Tschirschwitz and Rodehorst \cite{tschirschwitz2025}. We reject the assumption of uniform error. Instead, we propose a Dual-Layer Statistical Framework governed by a strict \textit{Data → Fit → Test} loop: 

\begin{enumerate} 
    \item \textbf{Parametric Drivers:} For geometric and topological errors, we fit empirical residuals to continuous distributions (\eg, Student's t, Von Mises, Beta), validating the fit via Kolmogorov-Smirnov (KS) tests and the Akaike Information Criterion (AIC). 
    \item \textbf{Validated Proxies:} For semantic and visual ambiguity (Category Mistakes, False Positives), we utilize foundation models (CLIP \cite{clip}, OWL-ViT \cite{owl}) as sampling distributions, validated against human confusion matrices using Mantel tests. 
\end{enumerate}

\textbf{Data Sources.} The generator is parameterized using pairwise comparisons extracted from the TexBiG \cite{tschirschwitz2022}, LVIS \cite{lvis}, and VinDr-CXR \cite{nguyen2022a} datasets. While our experimental validation utilizes a generalized distribution averaged across these sources, the parameters can be estimated independently to create dataset-specific generators.

\textbf{Section Structure.} The remainder of this section details the construction of the generator: 
\begin{enumerate} 
    \item We begin by reviewing prior noise generation models, highlighting the methodological leap from heuristics to empirical modeling (\cref{sec:prior_work}). 
    \item We describe the application of the error extraction pipeline \cite{tschirschwitz2025}, which isolates the raw disagreement data necessary for modeling (\cref{sec:error_extraction}) see also \cref{fig:noise_generator}.1. 
    \item We detail the core of our framework, systematically modeling each error type through our dual-layer approach (\cref{sec:modeling}) as shown in \cref{fig:noise_generator}.2. 
    \item Finally, we present the generator composition (\cref{fig:noise_generator}.3), validating the pipeline by quantifying the signal loss (saturation) that occurs when high-priority topological errors consume candidates for lower-priority modifications (\cref{sec:composition}).
\end{enumerate}

The full details of the generator can be found in the code: \href{https://github.com/Madave94/empirical-vision-noise-generator}{https://github.com/Madave94/empirical-vision-noise-generator}

\input{sec/synthetic_noise_table}

\subsection{Review of Prior Noise Generation Models}
\label{sec:prior_work}

Expanding upon \cref{sec:related_work_synthetic_noise_generation}, we provide a detailed comparison of previous synthetic noise generation models in \cref{tab:noise_comparison}. A general trend indicates that most methods for object detection remain relatively simplistic; notably, they often neglect complex error types such as combination and fragmentation, despite acknowledging their presence in real-world data \cite{chachula2023}. The most comprehensive approach found in the literature is the Corpus Shuffling Tool (CST) proposed by Mathet \etal \cite{mathet2012}. Although designed for textual data, its structural rigor serves as the foundation for our design.

Several assumptions are shared across multiple studies: (i) localization noise, specifically translation and scaling, is dependent on object size; (ii) errors follow uniform or normal distributions (\eg class permutations, translation magnitude/direction); and (iii) instance deletion (False Negatives) is modeled as an independent probabilistic event, unrelated to instance properties such as size or class. While individual heuristics may incorporate additional specific hypotheses, these three represent the core consensus across generation processes. A detailed breakdown is provided in \cref{tab:noise_comparison}.

\subsection{Recapping Error Extraction}
\label{sec:error_extraction}

Tschirschwitz and Rodehorst \cite{tschirschwitz2025} utilize the FiftyOne framework \cite{moore2020fiftyone} to analyze annotation variation, identifying four error types that we map to Mathet's taxonomy \cite{mathet2012}:
\begin{itemize}
    \item Bounding Box Variation $\rightarrow$ Localization Shift
    \item Overlooked/Missed Instance $\rightarrow$ Unmatched Instance (FP/FN)
    \item Mismatched Class $\rightarrow$ Category Mistake
    \item Merged/Unmerged Instance $\rightarrow$ Fragmentation/Combination
\end{itemize}
The extraction pipeline operates on pairwise annotator matching, simplifying the model by ignoring higher-order interdependencies. Due to the lack of ground truth, errors such as unmatched instances are treated symmetrically; a discrepancy between annotators implies ambiguity regarding whether an instance is a false positive or false negative. The same happens for fragmentation and combination.

\begin{figure}[t]
    \centering
    \includegraphics[width=1.0\linewidth]{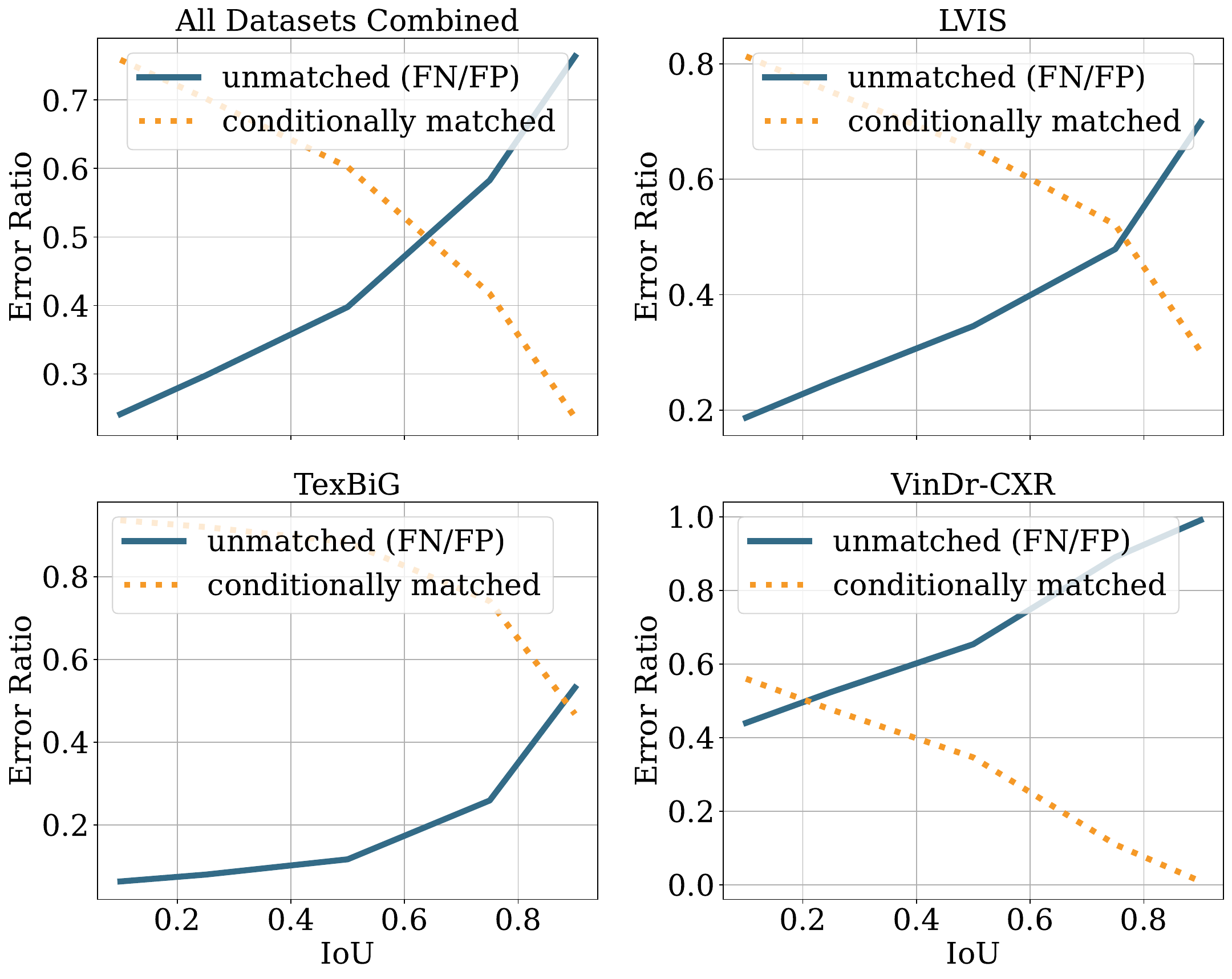}
    \caption{Impact of IoU threshold on error distribution. The plots illustrate the trade-off between unmatched instances (FP/FN) and conditionally matched instances (localization shifts, classification mistakes, combinations, and fragmentations) across the Combined, LVIS, TexBiG, and VinDr-CXR datasets.}
    \label{fig:equilibrium_points}
\end{figure}

To account for matching sensitivity, we evaluate multiple IoU thresholds: 0.1, 0.25, 0.5, 0.75, and 0.9. As illustrated in \cref{fig:equilibrium_points}, the ``equilibrium point'' where unmatched errors surpass conditionally matched errors varies significantly by dataset (\eg, $\approx 0.75$ for LVIS vs. $\approx 0.25$ for VinDr-CXR). Rather than fitting a variable threshold per dataset, we maintain a fixed 0.5 threshold to ensure a standardized definition of ``error'' across our noise generator. This value aligns with the minimum matching criteria for mAP evaluation in both LVIS \cite{lvis} and TexBiG \cite{tschirschwitz2022}. While VinDr-CXR \cite{nguyen2022a} utilizes a more lenient IoU of 0.4 for correct matches, we adhere to the 0.5 threshold to maintain parsimony and model consistency.

\subsection{Noise Type-Specific Modeling}
\label{sec:modeling}

\textbf{Note:} To ensure scale invariance across images of varying resolutions, all geometric modeling is performed using relative coordinates normalized to the range $[0, 1]$ rather than absolute pixel values.

When we refer to reference annotations, we mean the annotations from which the generator derives the noisy annotation, usually a human annotated instance.

\subsubsection{Localization Shift (\Deltaloc)} 
\label{sec:loc_shift}

We classify localization shift as a \textit{Parametric Driver}, a continuous error mode governing the geometric precision of the annotation. Unlike heuristic approaches that assume uniform noise, we model human spatial error as a superposition of size-dependent trends and heavy-tailed stochastic residuals. We decompose this shift into three statistically independent components: translation magnitude, directional bias, and scale variation.

\textbf{1. Translation Magnitude (Heavy-Tailed Residuals).} 
Standard noise generators often assume translation error is normally distributed. Our empirical analysis rejects this assumption. While translation magnitude correlates linearly with object size, the residuals exhibit significant heavy tails ($KS_{Student-t} < KS_{Normal}$), indicating that extreme outliers occur more frequently than Gaussian models predict. Comparison of Log-Likelihood fits confirms that a Student's $t$-distribution provides a superior fit to the residuals. Consequently, we model the total translation magnitude $\Delta_{trans}$ as:

\begin{equation}
\label{eq:translation}
\Delta_{trans} = \left| (\alpha_t + \beta_t \cdot A_{avg}) + \text{Clip}(\mathcal{T}_{\nu}(\mu, \sigma), q_{0.001}, q_{0.999}) \right|
\end{equation}

where $A_{avg}$ represents the average area of the paired instances, and $\alpha_t, \beta_t$ are coefficients derived from Multiple Linear Regression (MLR). $\mathcal{T}_{\nu}$ represents the random jitter drawn from the fitted Student's $t$-distribution. To prevent degenerate geometries during generation, we apply \textit{Winsorizing}, strictly clipping the stochastic residuals to the $[0.001, 0.999]$ quantiles of the fitted distribution.

\begin{figure}[t]
\centering
\includegraphics[width=1.0\linewidth]{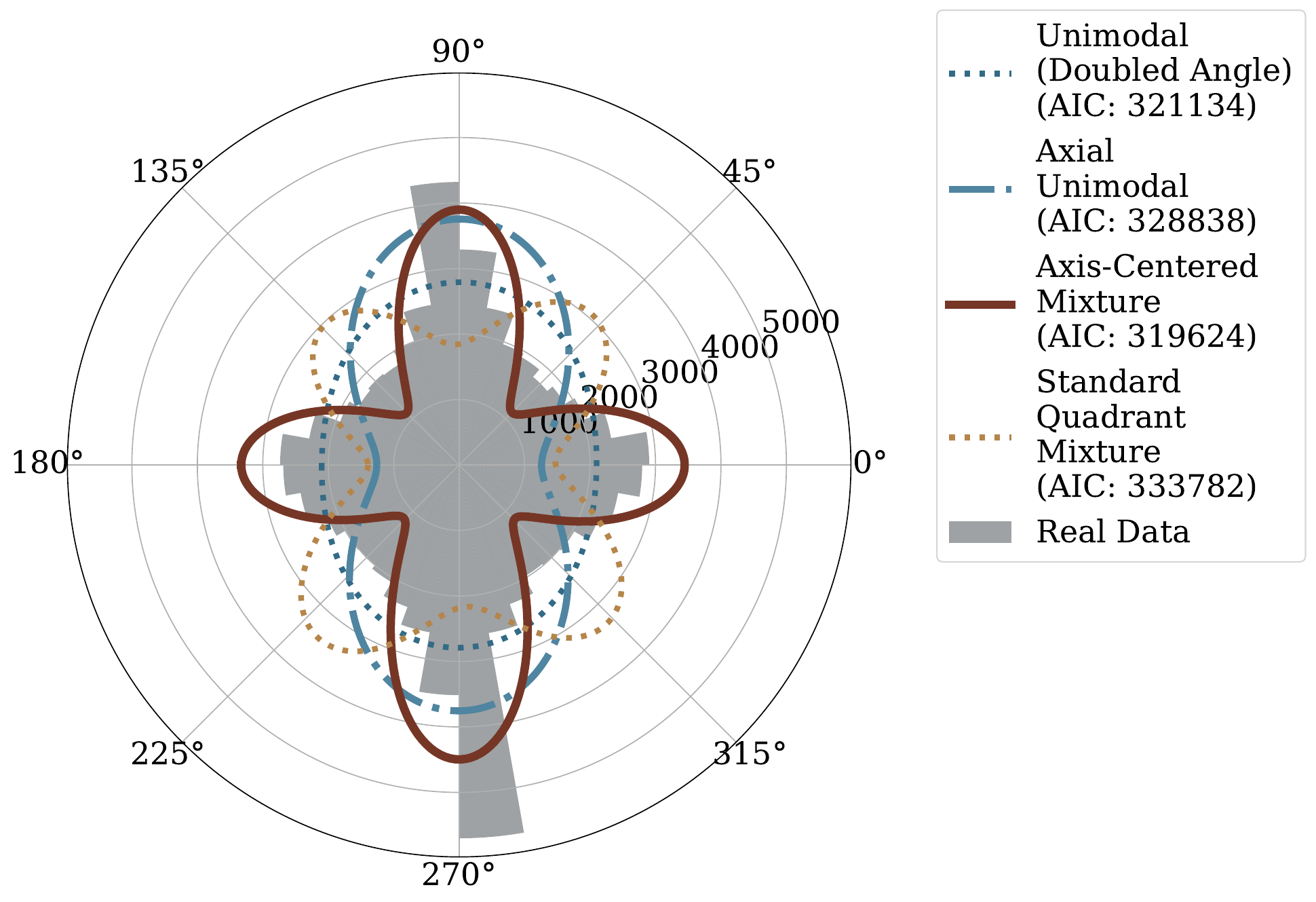}
\caption{Analysis of directional bias in bounding box shifts. The gray histogram represents the empirical distribution of shift angles. We tested four parametric models: Unimodal, Axial Unimodal, Axis-Centered Mixture, and Standard Quadrant Mixture. The \textbf{Axis-Centered Mixture Model} (brown line) yields the lowest Akaike Information Criterion (AIC $\approx$ 319,623), validating the hypothesis that human error is biased toward cardinal axes ($0^\circ, 90^\circ, 180^\circ, 270^\circ$).}
\label{fig:direction_bias}
\end{figure}

\textbf{2. Directional Bias (The Cardinal Hypothesis).} 
We scrutinize the common assumption that spatial error is isotropic (uniformly distributed in all directions). Hypothesis testing using both the Rayleigh test and the Kuiper test significantly rejects the null hypothesis of uniformity ($p < 0.001$) across all datasets.

\begin{figure*}[b]
    \centering
    \includegraphics[width=1.0\linewidth]{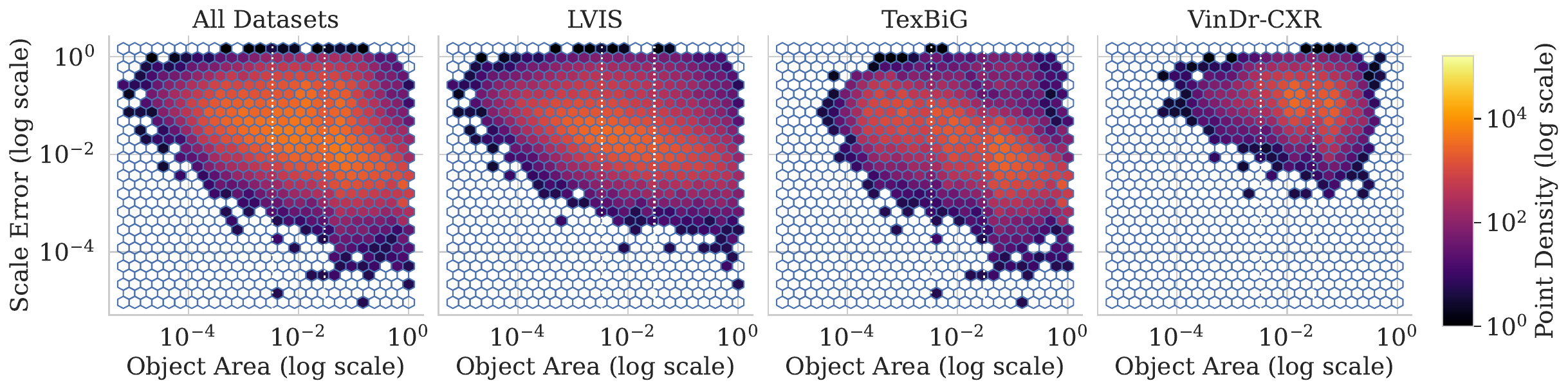}
    \caption{Relationship between relative scale error and object size. The hexbin plots visualize the density of scale errors across four datasets. A clear trend is visible where smaller objects (left side of x-axis) exhibit larger relative size differences (top of the y-axis), confirming the need for size-dependent noise modeling. The (thin) dashed white vertical lines divide into the COCO object size categories (adjusted to actual image size): Small ($<32^2$), Medium ($32^2$ to $96^2$), and Large ($>96^2$) for an image of $640 \times 480$.}
    \label{fig:scale_vs_size}
\end{figure*}

As illustrated in \cref{fig:direction_bias}, human error exhibits a strong ``Cardinal Bias,'' where annotators predominantly shift bounding boxes along the vertical and horizontal axes rather than diagonally. We model this probability density $P(\theta)$ using an Axis-Centered Mixture Model, defined as a weighted sum of four von Mises distributions centered at the cardinal directions:

\begin{equation}
\label{eq:direction}
P(\theta) = \sum_{k=1}^{4} w_k \cdot \text{VM}(\theta; \mu_k, \kappa_k)
\end{equation}

where $\mu_k \in \{0, \frac{\pi}{2}, \pi, \frac{3\pi}{2}\}$ and the concentration parameters $\kappa_k$ and weights $w_k$ are estimated via Maximum Likelihood Estimation (MLE).

\textbf{3. Scale Variation (Log-Space Asymmetry).} 
Scale errors are multiplicative rather than additive; a 10-pixel error is negligible for a large object but catastrophic for a small one. We model width ($s_w$) and height ($s_h$) scaling as independent drivers in log-space. Similar to translation, we observe a size-dependent trend where smaller objects suffer larger relative scale errors (see \cref{fig:scale_vs_size}).

We fit the absolute log-ratio of the reference and noisy dimensions to a linear model with Student's $t$ residuals. For the width dimension, this is defined as:

\begin{equation}
\label{eq:scale}
\ln(s_w) = (\alpha_w + \beta_w \cdot A_{avg}) + \mathcal{T}_{\nu_w}
\end{equation}

For general localization noise, the sign of the scaling factor is flipped with $p=0.5$ to ensure symmetric expansion and contraction.

\textbf{Engineering Constraints.} 
To transition from theoretical distributions to a valid generator, we enforce two strict engineering constraints. First, as noted in \cref{eq:translation}, the heavy tails of the Student's $t$-distribution require Winsorizing to prevent physically impossible outlier generation. Second, we enforce a boundary clip: the centroid of any generated instance must remain within the image coordinates. If a stochastic update violates this condition, the vector is clamped to the image edge, ensuring valid topology for downstream tasks.

    
\begin{figure*}
    \centering
    \includegraphics[width=1.0\linewidth]{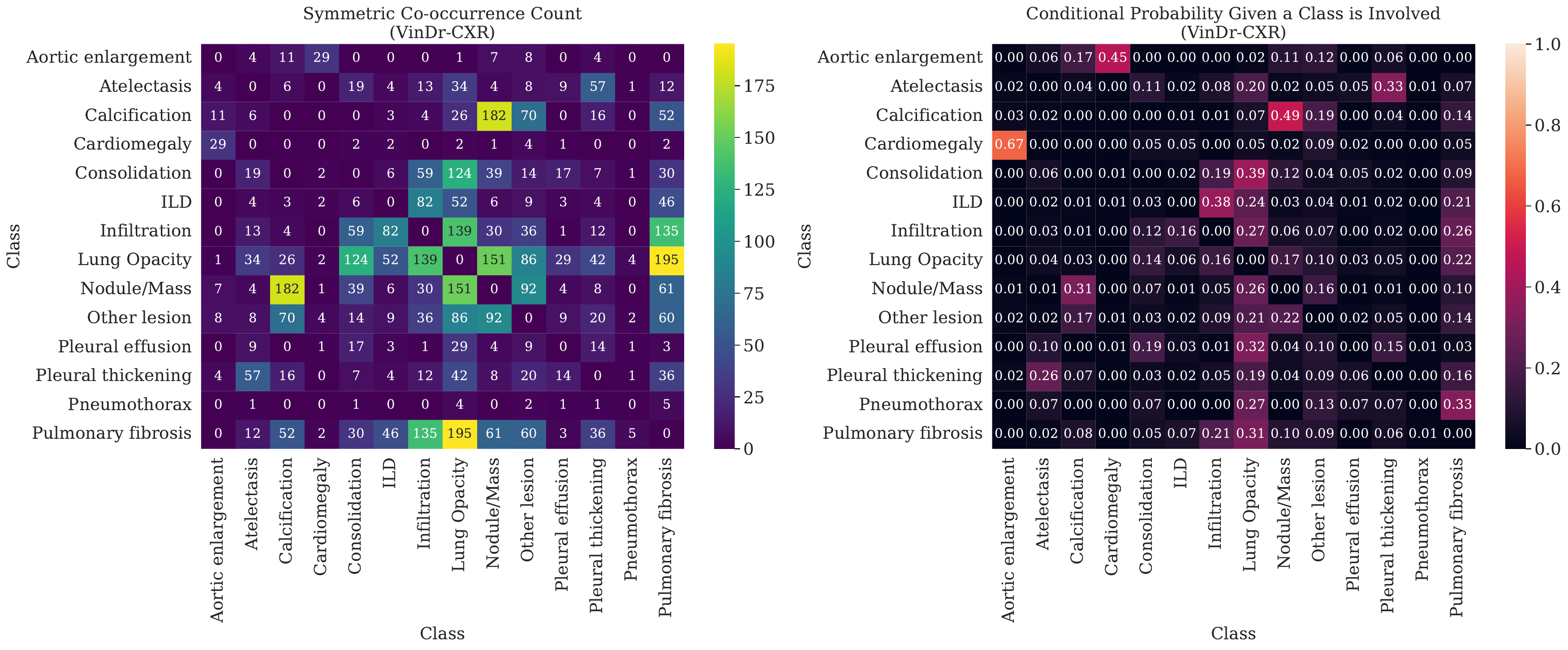}
    \caption{Analysis of classification co-occurrence for VinDr-CXR. Left: The symmetric (empirical) count of class confusions between annotators. Right: The conditional probability matrix, highlighting that confusion is often clustered among specific, semantically related pathologies (\eg, Pulmonary fibrosis vs. Lung Opacity).}
    \label{fig:co-occurance-vindr-cxr}
\end{figure*}

\subsubsection{Category Mistake (\Deltacls)}
\label{sec:cat_mistake}

We model category mistakes as a hybrid error, integrating both the \textit{Validated Proxy} and \textit{Parametric Driver} strategies. While the decision of \textit{which} class to substitute is driven by semantic proxies (visual embeddings), the physical realization of that error (the bounding box) is governed by re-calibrated parametric drivers.

\textbf{1. The Semantic Transition Matrix (Validated Proxy).}
Standard noise generators often assume class confusion is uniform (random permutation). We reject this hypothesis. A non-parametric Chi-squared permutation test on our data significantly rejects the null hypothesis of independence ($p < 0.05$). However, constructing robust empirical confusion matrices for long-tail datasets like LVIS is impossible due to data sparsity.

To address this, we utilize the visual embedding space of CLIP \cite{clip} as a validated proxy for human ambiguity. We compute a "Semantic Transition Matrix" based on the cosine similarity between the average visual centroids of all classes. We validate this proxy using a Mantel Test, which correlates the theoretical semantic similarity with the sparse empirical confusion matrix observed in the data. For LVIS, we observe a highly significant correlation ($r = -0.135, p = 0.0002$), confirming that semantic distance is a robust predictor of human labeling error.

In the generator, the probability of swapping the reference class $c_{ref}$ with a new class $c_{new}$ is modeled using a Softmax function over these similarities:

\begin{equation}
\label{eq:semantic_prob}
P(c_{new} | c_{ref}) = \frac{\exp(\text{sim}(\mathbf{v}_{c_{ref}}, \mathbf{v}_{c_{new}}) / \tau)}{\sum_{c' \in \mathcal{C}_{top10}} \exp(\text{sim}(\mathbf{v}_{c_{ref}}, \mathbf{v}_{c'}) / \tau)}
\end{equation}

where $\mathbf{v}_c$ is the average CLIP embedding for class $c$, and $\text{sim}(\cdot)$ is the cosine similarity. We introduce a semantic temperature $\tau=0.1$ to control the sharpness of the distribution, ensuring that confusions remain within a plausible semantic neighborhood (see \cref{fig:co-occurance-vindr-cxr}).

\textbf{2. Associated Localization Shift (Parametric Driver).}
Does a mislabeled object exhibit the same spatial error as a correctly labeled one? Our analysis suggests otherwise. We model the localization shift for misclassified instances using the same functional forms as \cref{sec:loc_shift} (Translation/Scale), but with re-estimated parameters.

Crucially, the directional bias changes. While standard localization error follows a 4-peak "Cardinal" distribution (snapping to x/y axes), misclassification errors are best fit by a Unimodal (Doubled Angle) von Mises distribution ($AIC \approx 11,752$) rather than the Axis-Centered Mixture ($AIC \approx 11,836$). This indicates that when annotators misidentify an object, their spatial precision degrades into a bi-modal axial distribution rather than the strict 4-way snapping observed in correct annotations.

\textbf{3. Occurrence Rate.}
The decision to trigger a category mistake is modeled as a global Bernoulli trial, independent of object size, parameterized by the empirical pairwise error rate calculated across all datasets ($p_{global} \approx 0.026$).


\subsubsection{Unmatched Instances (\Deltaum)}
\label{sec:unmatched}

We model unmatched instances (errors of existence) as a hybrid error. While the decision of where to place a hallucinated object is driven by visual proxies (OWL-ViT \cite{owl}), the governing laws of how many and which sizes are affected are strictly parametric. Due to the lack of ground truth, we treat FP and FN symmetrically; a disagreement implies ambiguity regarding existence, not a definitive error by one party.

\begin{figure}[t]
    \centering
    \includegraphics[width=1.0\linewidth]{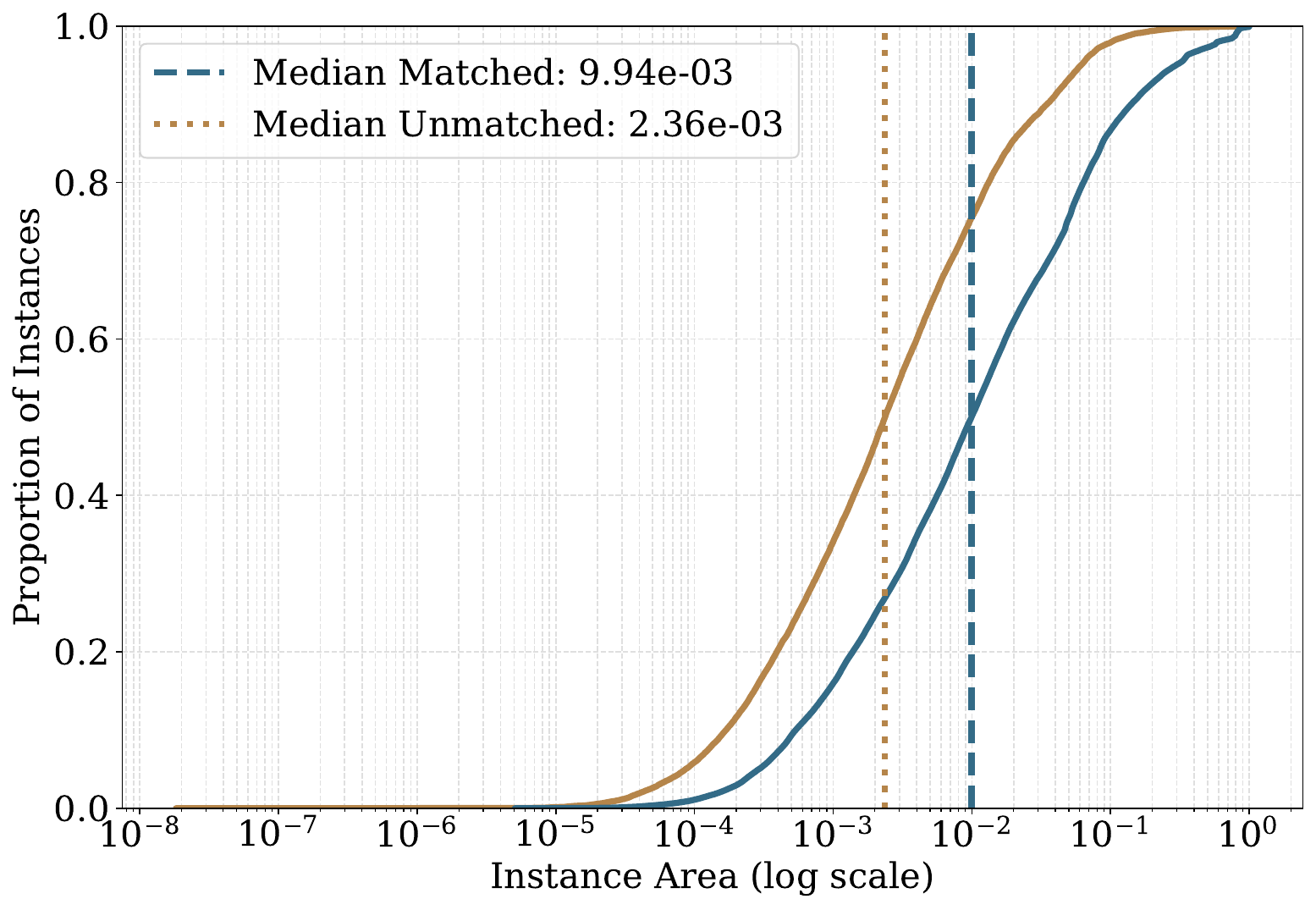}
    \caption{Cumulative Distribution Function (CDF) of instance areas for matched (blue) versus unmatched (orange) instances across all datasets. The distinct leftward shift of the unmatched curve indicates that missed or spurious annotations are systematically smaller than agreed-upon annotations.}
    \label{fig:unmatched_cdf}
\end{figure}

\textbf{1. Occurrence Rate (Parametric Driver).}
We hypothesize that the frequency of unmatched instances correlates with scene complexity. To model this, we fit a Bayesian Poisson regression to the conditional count of unmatched instances given the total annotation density ($|Y_i^r|$) of the rater. The expected number of noise events $\lambda_{um}$ is modeled as:

\begin{equation}
\label{eq:unmatched_rate}
\lambda_{um} = \exp(\alpha_{rate} + \beta_{rate} \cdot |Y_i^r|)
\end{equation}

where $\alpha_{rate}$ and $\beta_{rate}$ are the intercept and slope coefficients estimated via MCMC (Markov Chain Monte Carlo) sampling. Analysis of the posterior distributions ($\hat{R} \approx 1.0$) reveals distinct dataset characteristics: medical images (VinDr-CXR) show a steep error dependency ($\beta \approx 0.256$), while general scenes (LVIS) are more stable ($\beta \approx 0.021$). The final number of error events $k$ is drawn from $Poisson(\lambda_{um})$.

\textbf{2. Susceptibility and Pattern (Parametric Driver).}
Do annotators miss objects randomly? Our analysis rejects this heuristic. As shown in \cref{fig:unmatched_cdf}, unmatched instances are systematically smaller than matched instances. A Mann-Whitney U test confirms this size dependency is statistically significant ($p < 0.001$), proving that smaller objects are disproportionately prone to disagreement.

To capture this behavior, we fit a Logistic Regression on the log-transformed area of the instances. This yields a selection probability $P_{select}$ for any given candidate (whether existing or proposed):

\begin{equation}
\label{eq:selection_prob}
P_{select}(y) = \sigma(\alpha_{pat} + \beta_{pat} \cdot \ln(A_y))
\end{equation}

\textbf{3. Visual Ghosts (Validated Proxy).}
For each of the $k$ generated events, the system flips a coin ($p=0.5$) to decide between deletion (FN) or addition (FP):
\begin{itemize}
    \item \textbf{Deletion (FN):} Existing annotations are sampled for removal weighted by $P_{select}$ (Eq. \ref{eq:selection_prob}), ensuring that smaller objects are more likely to be "missed."
    \item \textbf{Addition (FP):} To generate realistic FP's, we cannot simply place random boxes. We utilize the open-vocabulary detector OWL-ViT \cite{owl} to generate a pool of visually plausible "ghost" proposals. We filter this pool to ensure validity: candidates must have an IoU $< 0.1$ with any existing annotation (to avoid creating duplicates). From this valid pool, a False Positive is sampled based on the size-weighted probability $P_{select}$. Finally, the class of the new instance is assigned using the Semantic Transition Matrix (CLIP \cite{clip} similarity) described in \cref{sec:cat_mistake}, ensuring the hallucinated object is semantically consistent with the scene context.
\end{itemize}

\begin{figure}[t]
    \centering
    \includegraphics[width=1.0\linewidth]{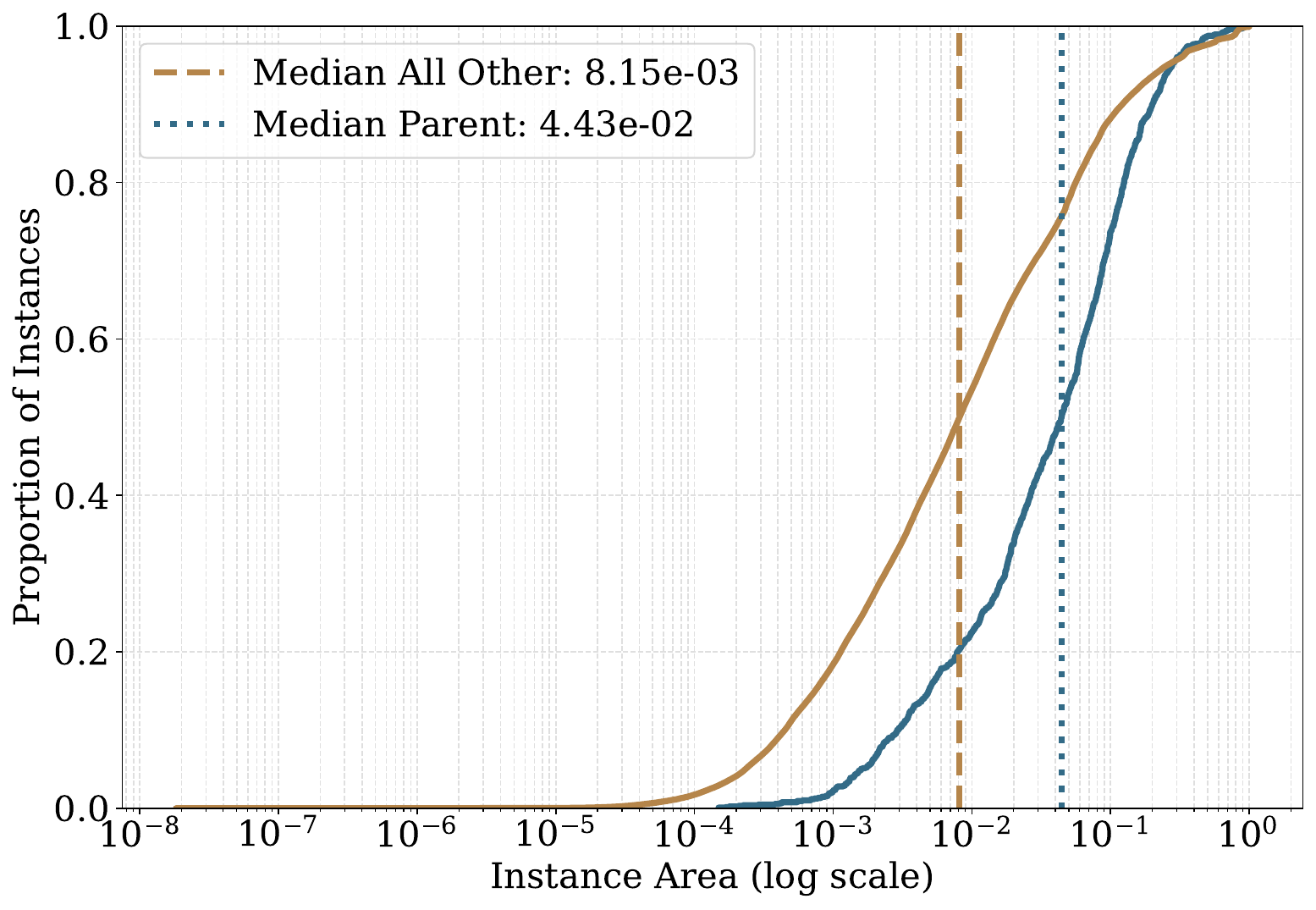}
    \caption{Cumulative Distribution Function (CDF) comparing the size of "parent" (blue) instances (involved in fragmentation/combination) versus all other instances. Parent instances are systematically larger, confirming that larger objects are more prone to split/merge errors.}
    \label{fig:mu_parent_size}
\end{figure}

\subsubsection{Fragmentation and Combination (\Deltatopo)}
\label{sec:frag_comb}

We classify topological errors (splitting one instance into two, or merging two into one) as a \textit{Reversible Parametric Driver}. We utilize a single geometric model to govern both directions: it is used generatively to split parents into fragments and discriminatively to identify valid merge candidates via maximum likelihood.

\textbf{1. Susceptibility (The Size Bias).}
Which objects are prone to topological disagreement? We hypothesize that complexity scales with size. A Mann-Whitney U test confirms that "parent" instances (those involved in split/merge errors) are statistically significantly larger than the general population ($p < 0.001$, see \cref{fig:mu_parent_size}). We model this susceptibility $P_{parent}$ using a Logistic Regression on the log-transformed area:

\begin{equation}
\label{eq:frag_susceptibility}
P_{parent}(y) = \sigma(\alpha_{frag} + \beta_{frag} \cdot \ln(A_y))
\end{equation}

\begin{figure}[b]
    \centering
    \includegraphics[width=1.0\linewidth]{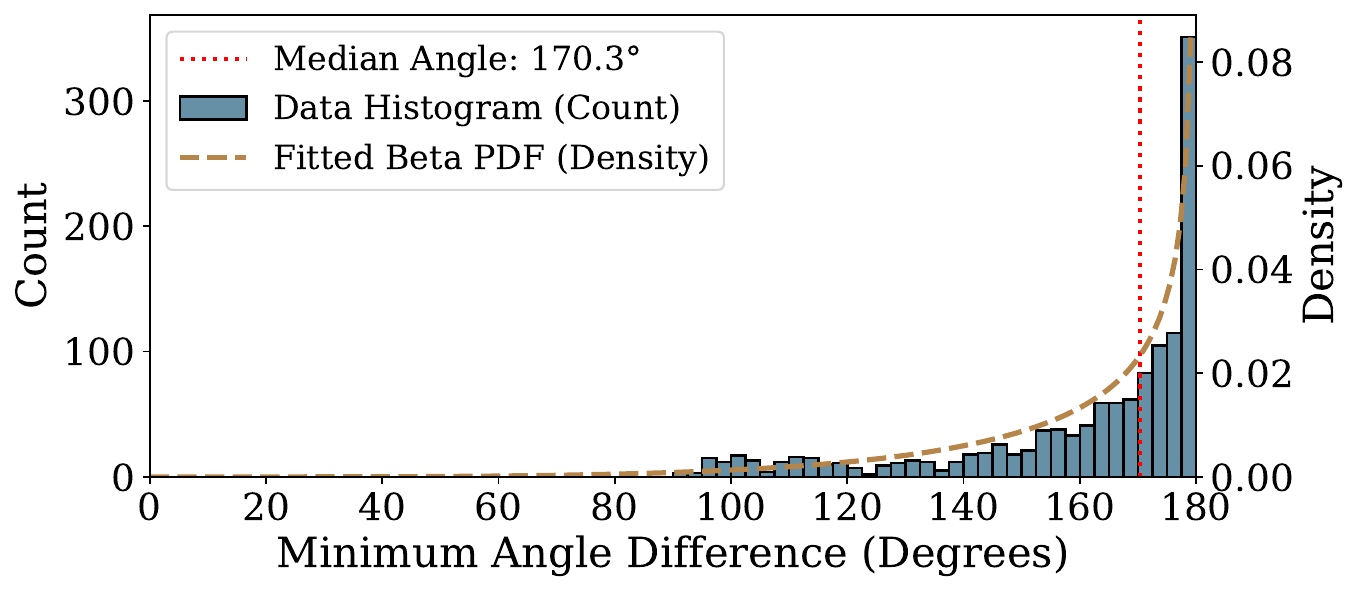}
    \caption{Distribution of the angular difference between two child fragments relative to their parent centroid. The peak near $180^\circ$ indicates that fragments tend to be located on opposing sides of the object, rather than clustered together.}
    \label{fig:child_angle_distribution}
\end{figure}

\textbf{2. The Geometry of Rupture (Generative).}
Once a parent is selected for fragmentation, we generate two children using a two-step dependency chain.
\begin{enumerate}[label=(\alph*)]
    \item \textbf{Parent-to-Child (Asymmetric):} The first child is generated using a modified localization model. Unlike standard shifts, scaling parameters are conditioned on the \textit{parent's} area (not average area) and are strictly negative (reduction). 
    
    \textit{Calibration Note:} Validation against empirical data revealed that Maximum Likelihood Estimation systematically underestimated child sizes (Median area ratio: $0.059$ synthetic vs. $0.088$ real). To correct this, we apply a post-hoc empirical calibration: a scalar factor ($\kappa \approx 0.20$) is added to the log-scaling intercepts, derived from the square root of the median discrepancy, ensuring synthetic fragments sum to a realistic proportion of the parent.

    \item \textbf{Child-to-Child (The "Opposing Sides" Hypothesis):} The second child is positioned relative to the first. We hypothesize that fragments typically represent opposing parts of an object (\eg head vs. torso). Empirical analysis confirms that 87.3\% of fragment pairs reside on opposing sides of the parent's major axis. We model the angular difference $\theta_{cc}$ between the two child vectors using a Beta Distribution:
\end{enumerate}

\begin{equation}
\label{eq:beta_angle}
\theta_{cc} \sim \text{Beta}(\alpha_{\theta}=4.53, \beta_{\theta}=0.53)
\end{equation}

As shown in \cref{fig:child_angle_distribution}, the distribution peaks near $180^\circ$ (normalized to 1.0), strictly enforcing the spatial separation observed in real disagreement.

\textbf{3. Combination via Model Inversion (Discriminative).}
To simulate combination (merging two instances), we invert the generative logic. Rather than using heuristics, we calculate a Geometric Likelihood Score $S(y_a, y_b)$ for every pair of same-class annotations. This score represents the joint probability density that the two instances \textit{could have been generated} as fragments of the same parent:

\begin{equation}
\label{eq:comb_score}
S(y_a, y_b) = f_t(\Delta_{trans}) \cdot f_w(s_w) \cdot f_h(s_h) \cdot f_{\theta}(\theta_{rel})
\end{equation}

where $f(\cdot)$ are the Probability Density Functions (PDF) of the fitted Student's $t$ (translation/scale) and Beta (angle) distributions. Pairs with high likelihood scores are merged into a single bounding box, ensuring that synthetic merges respect the statistical properties of human annotation patterns.

\subsection{Generator Composition}
\label{sec:composition}

\begin{figure*}[t] 
    \centering
    \begin{subfigure}[b]{0.32\textwidth}
        \centering
        \includegraphics[width=\linewidth]{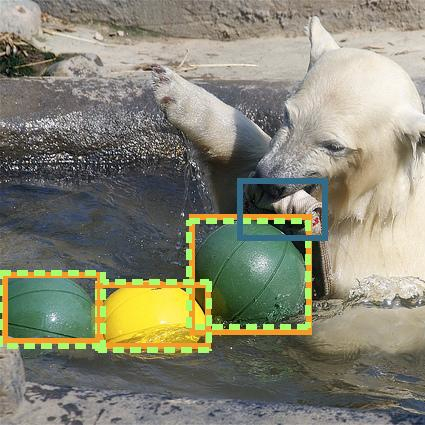}
        \label{fig:synth_vis_a}
    \end{subfigure}
    \hfill 
    \begin{subfigure}[b]{0.32\textwidth}
        \centering
        \includegraphics[width=\linewidth]{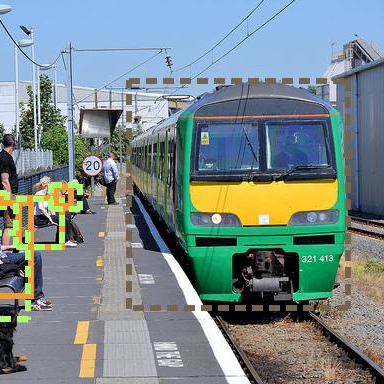}
        \label{fig:synth_vis_b}
    \end{subfigure}
    \hfill 
    \begin{subfigure}[b]{0.32\textwidth}
        \centering
        \includegraphics[width=\linewidth]{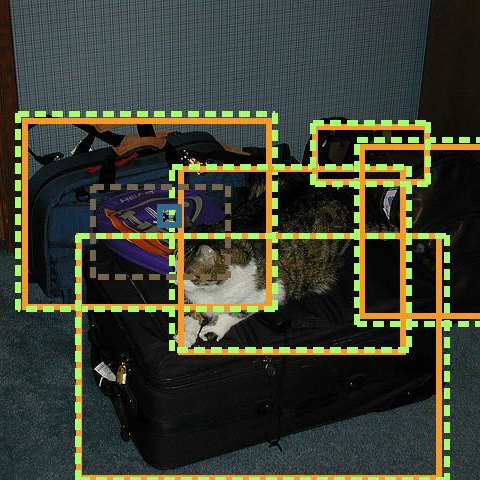}
        \label{fig:synth_vis_c}
    \end{subfigure}
    
    \caption{Visual comparison of synthetic (solid lines) and reference (dashed lines) annotations in LVIS. Matched pairs are shown in light orange (synthetic) and green (reference). Unmatched synthetic annotations (false positives) are navy, while unmatched reference annotations (false negatives) are brown. Images are cropped to a quadratic format.}
    \label{fig:synth_vis}
\end{figure*}


We combine the isolated parametric drivers and validated proxies into a unified \textbf{Empirical Noise Generator}. To resolve logical conflicts, such as the impossibility of shifting a bounding box that has been deleted, we enforce a strict topological hierarchy. The generator operates sequentially on the set of clean annotations $Y$, prioritizing errors that alter existence and topology over those that alter attributes or precision.

\textbf{1. The Execution Hierarchy.}
The generation pipeline executes four mutually exclusive stages. We maintain a dynamic registry of valid candidates; as high-priority processes consume "clean" annotations, the pool available for subsequent steps diminishes.
\begin{enumerate}
    \item \textbf{Unmatched Instances (Priority 1: Existence).} The generator first determines the set of active instances. FN remove candidates from the pool, while FP inject new proposals directly into the completed set, bypassing subsequent modification stages to preserve their "ghost" status.
    \item \textbf{Fragmentation/Combination (Priority 2: Topology).} Remaining candidates are evaluated for split/merge events. Successful topological changes consume the original parents and produce new, geometrically distinct children.
    \item \textbf{Category Mistake (Priority 3: Semantics).} The remaining clean instances are sampled for class permutation. Crucially, as noted in \cref{sec:cat_mistake}, misclassified instances receive their own specific localization noise profile and are effectively removed from the standard localization pool.
    \item \textbf{Localization Shift (Priority 4: Geometry).} Finally, any annotation that has survived the preceding stages without modification is subjected to the baseline localization shift (translation and scale) defined in \cref{sec:loc_shift}.
\end{enumerate}

\textbf{2. Magnitude Control ($\lambda$).}
The global noise intensity is controlled by a scalar factor $\lambda$. This parameter scales both the frequency of discrete error events and the variance of continuous geometric errors.
\begin{itemize}
    \item \textbf{Rate Scaling:} For discrete events (\eg category mistake), the empirical probability $p_{base}$ is scaled such that $p_{final} = \min(1.0, p_{base} \cdot \lambda)$.
    \item \textbf{Geometric Scaling:} For continuous drivers, $\lambda$ acts as a multiplicative factor on the stochastic residuals drawn from the Student's $t$ distributions: $\Delta_{final} = \Delta_{fitted} \cdot \lambda$.
\end{itemize}

\textbf{3. Signal Loss (Cannibalization).}
A consequence of this hierarchy is \textit{Signal Loss}. As $\lambda$ increases, high-priority processes (like deletion or fragmentation) consume a larger proportion of the available candidates. We track a metric termed "cannibalization", which counts the number of times a lower-priority process attempts to select a candidate that no longer exists.

As shown in \cref{tab:signal_lost}, at $\lambda=1.0$ (empirical baseline), signal loss is minimal ($4.2\%$). However, at extreme magnitudes ($\lambda=5.0$), nearly $30\%$ of theoretically sampled errors are "cannibalized." This saturation is not a flaw but a valid simulation of information entropy: as noise overwhelms the signal, the distinctions between specific error types vanish, leaving only a chaotic distribution.

\begin{table}[h]
\centering
\small
\caption{Signal Loss Analysis (LVIS). As noise magnitude $\lambda$ increases, the competition for finite candidates leads to saturation, where the effective error rate diverges from the theoretical sampling rate.}
\label{tab:signal_lost}
\begin{tabular}{lccccc}
Magnitude $\lambda$ & 0.25 & 0.50 & 1.00 & 2.00 & 5.00 \\
\hline
Signal-loss ratio   & $0.2\%$ & $1.3\%$ & $4.2\%$ & $10.1\%$ & $28.3\%$ \\
\end{tabular}
\end{table}

\textbf{Limitations.} We identify three primary limitations of this framework. 
(1) \textit{Saturation at high magnitudes:} As $\lambda$ increases, the competition for unmodified candidates intensifies, leading to signal loss. At extreme magnitudes ($\lambda \gg 2.0$), the reference signal degrades until the output resembles random noise. This saturation is expected behavior, as noise levels significantly beyond the empirical baseline eventually obscure the underlying signal entirely.
(2) \textit{Visual independence:} With the exception of proposal based false positive generation, the stochastic processes are decoupled from image content. While factors such as occlusion or blur drive real world human error, explicit modeling of these visual cues was not required for the validation of agreement metrics. Future iterations targeting model training data augmentation would benefit from incorporating such uncertainty maps.
(3) \textit{Validation scope:} Our validation focuses on the goodness of fit between empirical data and our parametric models. While a direct statistical comparison between the generated synthetic distributions and the original empirical distributions remains a subject for future investigation, this framework nevertheless significantly advances the state of the art by modeling complex and interdependent error modalities rather than relying on isolated heuristics or black-box machine label noise.

%% file: sec/synthetic_noise_table.tex
\begin{table*}[t]
\caption{Comparison of Synthetic Noise Generation Methods. We use the taxonomy of Mathet \etal \cite{mathet2012} to model four channels: (i) \textbf{Loc.} (localization: shift, scale), (ii) \textbf{Unm.} (unmatched instances: missing, extra), (iii) \textbf{Cat.} (category mismatches), and (iv) \textbf{Fra./Com.} (fragmentation/combination). Without a reference, unmatched instances are sampled symmetrically. The study of Hu \etal \cite{hu2019} is excluded, since we didn't have any specifications of the generated noise. Note that Mathet \etal originally applied these concepts to text segmentation, not object detection, we present their idea for noise generation here as well since it's the most comprehensive.}
\label{tab:noise_comparison}
\centering
\small
\renewcommand{\arraystretch}{1.2}
\setlength{\tabcolsep}{4pt} 
\begin{tabularx}{\linewidth}{@{} p{0.14\linewidth} p{0.07\linewidth} >{\hsize=0.9\hsize}X >{\hsize=1.1\hsize}X @{}}
\toprule
\textbf{Study} & \textbf{Type} & \textbf{Modelling} & \textbf{Hypothesis (Assumptions)} \\ 
\midrule

\multirow{2}{=}{Agnew \etal \cite{agnew2023}} 
& Loc. & Uniform expansion (0-10 pixel buffering) of BBox & Noise is primarily expansion (outward shift). \\
& Loc. & Radial Gaussian noise ($1-5\mu$) on mask vertices & Vertex precision varies normally; errors localize to edges. \\ 
\midrule

\multirow{4}{=}{Mathet \etal \cite{mathet2012} (CST)} 
& Unm. & Random dropout \& addition (class is based on marginal dist.) & Errors are probabilistic; False Positives follow dataset statistics. \\
& Loc. & Random uniform shift & Shift magnitude is proportional to object length. \\
& Cat. & Confusion matrix (prevalence \& overlap based) & Errors correlate with class frequency and semantic overlap. \\
& Fra. & Randomly select element and split it & Annotators mistakenly split single entities. \\ 
\midrule

\multirow{1}{=}{Bär \etal \cite{bar2023}} 
& Loc. & Uniform noise ($\Delta h, \Delta w$) $\propto$ width/height & Error scales with object size; no directional bias. Same as \cite{li2020d} \\ 
\midrule

\multirow{3}{=}{Chachuła \etal \cite{chachula2023} (CLOD)} 
& Loc. & Random angle shift + scale change $\propto$ size & Errors involve position and size relative to dimensions. \\
& Unm. & Randomly adding spurious or deleting boxes & Presence/absence errors occur randomly. \\
& Cat. & Uniform label noise (random permutation) & Class confusion is random (no hierarchy assumed). \\ 
\midrule

\multirow{3}{=}{Li \etal \cite{li2020d}} 
& Loc. & Uniform noise ($\Delta h, \Delta w$) $\propto$ width/height & Error scales with object size; no directional bias. Same as \cite{bar2023}. \\
& Cat. & Symmetric noise (random flip) & Confusion probability is uniform across classes. \\
& Unm. & Proposals from a poorly trained detector (FP), this are machine generated label noise & FPs mimic realistic machine bias rather than random noise. \\ 
\midrule

\multirow{2}{=}{Liu \etal \cite{liu2022} (NLTE)} 
& Cat. & Uniform random substitution & Class confusion is random. \\
& Unm. & Deletion if substituted label is ``background'' (FN) & Missing objects are misclassified as background. \\ 
\midrule

\multirow{1}{=}{Liu \etal \cite{liu2022a} } 
& Loc. & Uniform dist. (10-40\%) relative to original bbox & Errors are strictly size-dependent relative shifts. \\ 
\midrule

\multirow{3}{=}{Chan \etal \cite{chan2023} (ODFI)} 
& Unm. & Random removal; ``Redundant'' (duplicate + offset) & FPs are often ``ghost'' boxes (slight shifts of true objects). \\
& Loc. & Random reposition + size reduction (30\%) & Errors involve deterministic shrinkage/displacement. \\
& Cat. & Superclass vs. Subclass swapping & Errors are hierarchically semantic (e.g. car$\leftrightarrow$truck $\ne$ car$\leftrightarrow$person). \\ 
\midrule

\multirow{3}{=}{Chadwick \& Newman \cite{chadwick2019}} 
& Cat. & Swap classes of nearby labels (Pair Noise) per image & Errors are systematic/spatially correlated (contextual confusion). \\
& Loc. & Shift and scale using Normal distribution & Spatial error is Gaussian; independent of object size. \\
& Unm. & Random addition (max 1/image) and removal & Spurious objects are rare events; missing is probabilistic. \\ 

\bottomrule
\end{tabularx}
\end{table*}

%% file: sec/cost_functions.tex
\section{Comparing Cost Functions}
\label{sec:cost_functions}

\begin{figure*}
    \centering
    \includegraphics[width=1.0\linewidth]{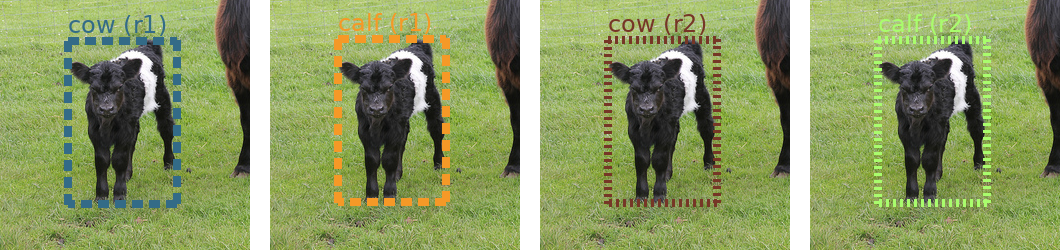}
    \caption{Rater-level annotations for LVIS image No. 467776 (shows only the region of interest) illustrating a conceptual edge case where class information must influence the correspondence cost. The two panels on the left show annotations from rater r1, and the two on the right from rater r2. Both raters provide two labels (cow and calf) for the very same physical instance, and both classes occupy effectively identical spatial regions. Because object detection allows overlapping instances, the spatial component alone provides no discriminative signal: all four boxes overlap nearly perfectly, and localization-based matching is therefore ambiguous. A class-aware cost function resolves this ambiguity in a principled way by preferring correspondences that preserve each rater’s intended class assignment rather than artificially swapping them. This example illustrates why incorporating class information into the matching cost is conceptually necessary for certain real-world annotation patterns, even before considering any quantitative evaluation.}
    \label{fig:cow_calf_figure}
\end{figure*}

\begin{figure*}
    \centering
    \includegraphics[width=1.0\linewidth]{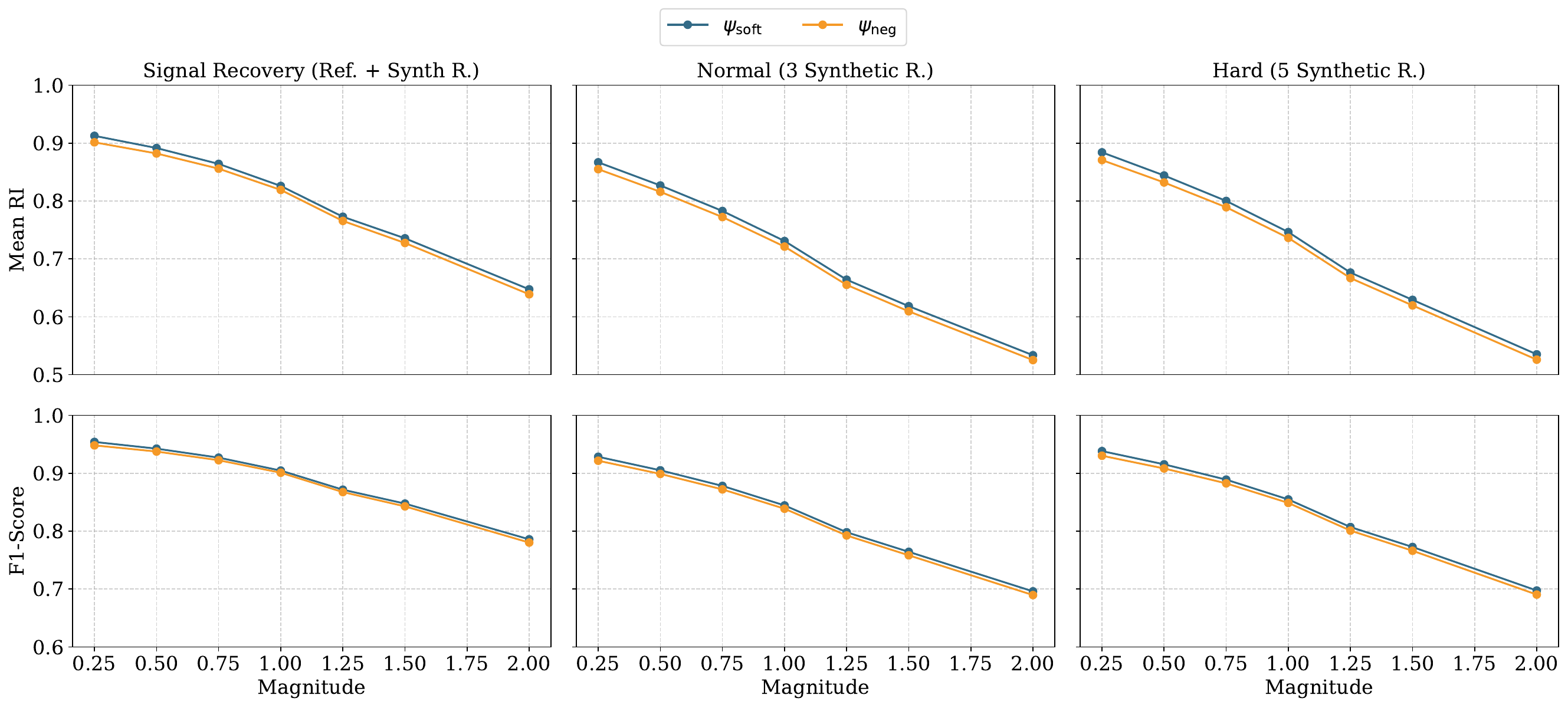}
    \caption{Comparison of the two correspondence cost functions on LVIS using the Greedy solver across noise magnitudes and rater configurations. The category-lenient cost $\psi_{\text{soft}}$ consistently achieves slightly higher filtered Rand Index and F1-scores than the baseline $\psi_{\text{neg}}$. While the differences are small, they are systematic and stable across all settings, indicating that incorporating class information into the cost improves correspondence robustness.}
    \label{fig:cost_function comparison}
\end{figure*}

To compare the effect of incorporating semantic information into the correspondence cost, we evaluate a baseline cost function that relies solely on localization agreement. This cost, denoted $\psi_{\text{neg}}$, assigns higher preference to pairs with greater spatial overlap but remains indifferent to class identity. Formally, it is defined as:
\begin{equation}
    \psi_{\text{neg}}(d_{\text{loc}},d_{\text{cls}}) = - d_{\text{loc}},
\end{equation}

The class aware cost function is in the main paper as \cref{eq:category_lenient}.

The qualitative example in \cref{fig:cow_calf_figure} illustrates a broader phenomenon frequently encountered in instance-based vision tasks: multiple semantic categories may legitimately occupy nearly identical spatial regions. When localization alone is used as the correspondence signal, such situations induce ambiguity; spatial overlap becomes insufficient to determine which annotations should be matched. In these cases, solvers that ignore class information are prone to semantically implausible correspondences, especially under realistic annotation noise.

\Cref{fig:cost_function comparison} quantitatively confirms this intuition. Across all noise magnitudes and rater configurations, the category-lenient cost $\psi_{\text{soft}}$ consistently outperforms the purely localization-based $\psi_{\text{neg}}$. While the numerical differences are modest, they are systematic and robust: $\psi_{\text{soft}}$ reliably avoids precision-degrading errors in which spatially similar but semantically incompatible annotations are merged. These findings show that even a simple class-consistency term stabilizes correspondence formation, particularly when spatial overlap is high due to class definitions that allow or encourage closely colocated instances.

%% file: sec/robustness_analysis.tex
\section{Robustness Analysis of Data-Driven Calibration}
\label{sec:robustness_analysis}

To evaluate the stability of the distance metric $d$ and calibration anchor $\tau^*$, we perform a non-parametric percentile bootstrap. We resample $N$ images with replacement over 100 iterations to compute 95\% confidence intervals. Because calibration is dataset-dependent, we apply this analysis to object detection on VinDr-CXR \cite{nguyen2022a}, TexBiG \cite{tschirschwitz2022}, and LVIS \cite{lvis} using IoU. As \cref{tab:robustness_1} demonstrates, $\tau^*$ remains highly stable across samples. This tight variance confirms that the empirical noise profile is a consistent property of the dataset, justifying minor threshold adjustments (e.g., rounding to align with standard IoU benchmarks) provided they fall near these confidence bounds.

\begin{table*}[tb]
\centering
\caption{Global robustness analysis of the calibration anchor ($\tau^*$) and KS statistic across 100 bootstrap iterations (All Sizes). CI denotes the 95\% Confidence Interval.}
\label{tab:robustness_1}
\begin{tabular}{lrrr}
\toprule
\textbf{Dataset} & \textbf{Annotations} & \textbf{$\tau^*$ Mean (95\% CI)} & \textbf{KS Mean (95\% CI)} \\
\midrule
LVIS \cite{lvis} & $100,480$ & $0.654$ ($[0.643, 0.668]$) & $0.798$ ($[0.786, 0.807]$) \\
TexBiG \cite{tschirschwitz2022} & $52,372$ & $0.435$ ($[0.420, 0.455]$) & $0.803$ ($[0.796, 0.810]$) \\
VinDr-CXR \cite{nguyen2022a} & $36,096$ & $0.996$ ($[0.996, 0.997]$) & $0.684$ ($[0.675, 0.692]$) \\
\bottomrule
\end{tabular}%
\end{table*}

Standard IoU correlates poorly with human perception, disproportionately penalizing localization errors in smaller objects \cite{strafforello2022,llerena2025}. To investigate this bias, we stratify our robustness analysis by relative instance size (small, medium, and large, scaled proportionally from standard COCO definitions to relative image area). We employ an asymmetric matching strategy: target instances of a specific size class are matched against the entire pool of reference annotations to prevent artificial boundary exclusion.

\cref{tab:robustness_2} reveals a pronounced scale dependency across all domains. Smaller objects consistently require significantly more lenient (higher) distance thresholds to successfully separate signal from noise, whereas larger objects demand much stricter tolerances. Consequently, selecting a single global similarity threshold imposes a fundamental trade-off: stricter thresholds accurately capture geometric precision but discard valid small-object annotations as noise, while lenient thresholds capture small objects but misclassify stochastic chance overlaps as true consensus for large objects.

\begin{table*}[b]
\centering
\caption{Size-stratified robustness analysis using asymmetric matching. Small objects require highly lenient (higher) distance thresholds compared to large objects, exposing the scale bias of IoU.}
\label{tab:robustness_2}
\begin{tabular}{llrrr}
\toprule
\textbf{Dataset} & \textbf{Size} & \textbf{Target Anns} & \textbf{$\tau^*$ Mean (95\% CI)} & \textbf{KS Mean (95\% CI)} \\
\midrule
\multirow{3}{*}{LVIS \cite{lvis}} 
 & Small & $35,574$ & $0.976$ ($[0.973, 0.978]$) & $0.773$ ($[0.750, 0.795]$) \\
 & Medium & $38,558$ & $0.748$ ($[0.725, 0.771]$) & $0.863$ ($[0.855, 0.870]$) \\
 & Large & $26,348$ & $0.488$ ($[0.473, 0.501]$) & $0.889$ ($[0.883, 0.894]$) \\
\midrule
\multirow{3}{*}{TexBiG \cite{tschirschwitz2022}} 
 & Small & $12,769$ & $0.695$ ($[0.644, 0.764]$) & $0.904$ ($[0.894, 0.918]$) \\
 & Medium & $15,376$ & $0.607$ ($[0.570, 0.651]$) & $0.880$ ($[0.870, 0.889]$) \\
 & Large & $24,227$ & $0.117$ ($[0.113, 0.121]$) & $0.843$ ($[0.834, 0.851]$) \\
\midrule
\multirow{3}{*}{VinDr-CXR \cite{nguyen2022a}} 
 & Small & $3,976$ & $0.995$ ($[0.995, 0.996]$) & $0.524$ ($[0.497, 0.553]$) \\
 & Medium & $17,262$ & $0.996$ ($[0.996, 0.997]$) & $0.659$ ($[0.649, 0.667]$) \\
 & Large & $14,858$ & $0.875$ ($[0.848, 0.956]$) & $0.776$ ($[0.768, 0.785]$) \\
\bottomrule
\end{tabular}%
\end{table*}

Crucially, this scale bias is an inherent flaw of the chosen distance metric (IoU), not \KalphaLOS{}. Because \KalphaLOS{} operates as a modular meta-algorithm, it successfully isolates and exposes this metric bias. Future applications can mitigate this issue entirely by substituting IoU with a scale-invariant distance metric within the \KalphaLOS{} framework.

%% file: sec/global_vs_mean.tex
\section{Global vs Mean \KripA{}}
\label{sec:global_vs_mean_alpha}

The primary \KalphaLOS{} metric is computed on a per-image basis, and the dataset-wide score is reported as the mean of these values (mean \KripA{}). We empirically validate this image-centric approach because it ensures that agreement is balanced across independent scenes.

An alternative approach is to concatenate all per-image reliability matrices and run a single calculation on the resulting global matrix. We refer to this unvalidated variant as the secondary global metric (global \KripA{}). While global \KripA{} can be useful for datasets with highly uniform instance counts per image or for cases where agreement on absence is intentionally weighted lower, we treat it strictly as a secondary diagnostic due to inherent statistical biases:

\begin{itemize}
    \item \textbf{Instance-Based Domination:} global \KripA{} weights the evaluation by instance count rather than by scene. A single dense image containing 100 instances will mathematically overwhelm the consensus of 100 sparse images containing one instance each. This disproportionately biases the metric toward the geometric precision of clustered objects (e.g., dense document layouts) while masking disagreement in sparse scenes.
    
    \item \textbf{Distortion of Expected Disagreement ($D_e$):} Concatenating matrices merges the class distributions of all independent scenes. The metric's baseline for chance agreement ($D_e$) becomes skewed by the most frequently annotated classes in the densest images, artificially altering the penalty for misclassification across the rest of the dataset.
    
    \item \textbf{Devaluation of Absence:} In domains like medical analysis, an entirely empty image represents a critical consensus on the absence of a finding. In mean \KripA{}, this perfect agreement holds equal weight to an annotated scan. In global \KripA{}, to mitigate the issue of the matrix dropping the empty image entirely, it is injected as a single virtual \texttt{NO\_OBJECT} agreement. While this reduces the issue it still remains. Consequently, an image with 50 annotations carries 50 times the mathematical weight of a clinically significant empty image.
\end{itemize}

%% file: sec/downstream_tasks.tex
\section{Downstream Tasks}
\label{sec:downstream_tasks}

Regardless of the task of computer vision, \KalphaLOS{} arrives at a nominal reliability matrix. Since the data structure is always the same, downstream diagnostics (\cref{fig:kalos}.4) become available. By re-computing $\alpha$ on filtered or permuted views of this matrix, the framework enables granular diagnostics beyond a single summary score.

\noindent\textbf{Per-Image Agreement Distribution:} Plotting the per-image $\alpha$ scores (see \cref{fig:per_image_agreement_distribution}) reveals the stability of the consensus. A left-skewed distribution identifies specific problematic images or ambiguous scenarios, enabling targeted refinement of annotation guidelines during dataset creation.

\noindent\textbf{Localization Sensitivity Analysis:} This analysis quantifies the agreement "lost" to boundary imprecision. By computing the delta between agreement at the calibration anchor ($\tau^*$) and a stricter threshold ($\tau_{strict}$), \ie, $\Delta = \alpha(\tau^*) - \alpha(\tau_{strict})$, we isolate true geometric jitter from fundamental existence disagreement.

\noindent\textbf{Class Recognition Difficulty:} We diagnose semantic ambiguity by filtering the reliability matrix for a single class $c$. Crucially, under the completeness assumption, any rater who failed to annotate $c$ is assigned $\text{\textsc{no\_object}}$. Re-computing $\alpha$ on this subset distinguishes well-defined classes (\eg, \texttt{person}: $\alpha \approx 0.8$) from ambiguous ones (\eg, \texttt{distant bird}: $\alpha \approx 0.4$).

\noindent\textbf{Collaboration Clusters:} This analysis identifies "schools of thought" by generating a pairwise vitality matrix. High pairwise agreement between subsets of raters reveals implicit conventions or institutional biases distinct from the global consensus.

\noindent\textbf{Intra-Annotator Agreement:} By treating a rater's annotations at times $t_0$ and $t_1$ as independent rows in the reliability matrix, \KalphaLOS{} quantifies self-consistency without architectural changes.

\noindent\textbf{Annotator Vitality:} Following Nassar \etal \cite{nassar2019}, we measure an individual rater's contribution to the consensus. The vitality $V_r$ for rater $r$ is defined as:
\begin{equation}
    V_r = \alpha_{R} - \alpha_{R \setminus \{r\}}
\end{equation}
where $\alpha_R$ is the score with the full cohort and $\alpha_{R \setminus \{r\}}$ is the score with rater $r$ removed. A positive $V_r$ identifies a consensus builder; a negative $V_r$ identifies a source of noise or deviation.

%% file: sec/kalos_applications.tex
\section{Application of \KalphaLOS{}}
\label{sec:application_of_kalos}

This section demonstrates the adaptability of the \KalphaLOS{} meta-algorithm across distinct computer vision domains. For each task, we derive the principled configuration via distributional analysis, identifying the optimal distance function ($d_{loc}$) and the calibration anchor ($\tau^*$) required to statistically separate annotator signal ($D_o$) from stochastic noise ($D_e$).

Crucially, we apply the completeness assumption globally for these experiments: we assume that raters annotate all instances they perceive. Consequently, if a rater fails to annotate a unit discovered by the group, it is encoded as an explicit active disagreement (\textsc{no\_object}), penalizing the score as a false negative, rather than treated as missing data. We present this diagnostic pipeline on three diverse tasks: (1) instance segmentation on TexBiG~\cite{tschirschwitz2022}, (2) 3D volumetric segmentation on LIDC-IDRI~\cite{armato2015lidc}, and (3) pose estimation on MARS~\cite{segalin2021}.

\textbf{Note:} For the distributional analysis distances $d$ are used. However, it is common practice in most tasks (object detection, instance segmentation) to use similarities instead. Hence, for the distributional analysis, we provide distance (and as a secondary axis similarity values), but in later analysis similarities are used in accordance with standard practice. The distance threshold is shown as $\tau$ and similarity thresholds as $\tau_s$.

\subsection{Instance Segmentation}
\label{sec:instance_segmentation}

TexBiG \cite{tschirschwitz2022} is a dataset on complex layouts of historical documents with dense annotations covering all elements in the layout beside the background (mostly white).

\textbf{Principled Configuration.} To adapt \KalphaLOS{} for instance segmentation, we first determine the optimal localization distance function $d_{loc}$ and the associated calibration anchor $\tau^*$. We evaluate three candidate functions: Polygon IoU, Mask GIoU, and L2 Centroid distance. As the framework requires a distance metric $d$, we normalize and invert similarity scores where necessary. Annotations are defined as $\tilde{y}_{ik}^{r}=(\tilde{p}^r_{ik},\tilde{c}^r_{ik})$, where $\tilde{p}^r_{ik}$ represents the segmentation polygon and $\tilde{c}^r_{ik}$ remains the class, same as for bounding boxes.

\begin{enumerate} 
    \item \textbf{Polygon IoU:} We calculate the standard intersection over union for polygonal masks. To convert this similarity into a distance, we invert it: \begin{equation} d_{\text{IoU}}(\tilde{p}_{ik}^{a}, \tilde{p}_{il}^{b}) = 1 - \frac{|\tilde{p}_{ik}^{a} \cap \tilde{p}_{il}^{b}|}{|\tilde{p}_{ik}^{a} \cup \tilde{p}_{il}^{b}|} \end{equation}
    \item \textbf{Mask GIoU:} The Generalized IoU (GIoU) accounts for spatial proximity in non-overlapping shapes. 
        \begin{equation}
        \operatorname{GIoU}(\tilde{p}_{ik}^{a}, \tilde{p}_{il}^{b}) = \operatorname{IoU}(\tilde{p}_{ik}^{a}, \tilde{p}_{il}^{b}) - \frac{|C \setminus (\tilde{p}_{ik}^{a} \cup \tilde{p}_{il}^{b})|}{|C|},
    \end{equation}
    where $C$ is the smallest convex hull enclosing both $\tilde{p}_{ik}^{a}$ and $\tilde{p}_{il}^{b}$. Since the standard GIoU range is $[-1, 1]$, we first normalize it to a similarity $S \in [1]$ and then invert it to define the distance:
    \begin{equation}
        d_{\text{GIoU}} = 1 - \frac{1 + \text{GIoU}}{2}
    \end{equation}
    
    \item \textbf{L2 Centroid Distance:} We compute the Euclidean distance between geometric centroids defined as $\mathbf{\tilde{p}^r_{ik}}$, normalized by the image diagonal to ensure the range $[0,1]$.
    \begin{equation}
        d_{L2}(\tilde{p}_{ik}^{a}, \tilde{p}_{il}^{b}) = \|\mathbf{m}(\tilde{p}_{ik}^{a}) - \mathbf{m}(\tilde{p}_{il}^{b})\|_2.
    \end{equation}
\end{enumerate}

\begin{figure}
    \centering
    \includegraphics[width=1.0\linewidth]{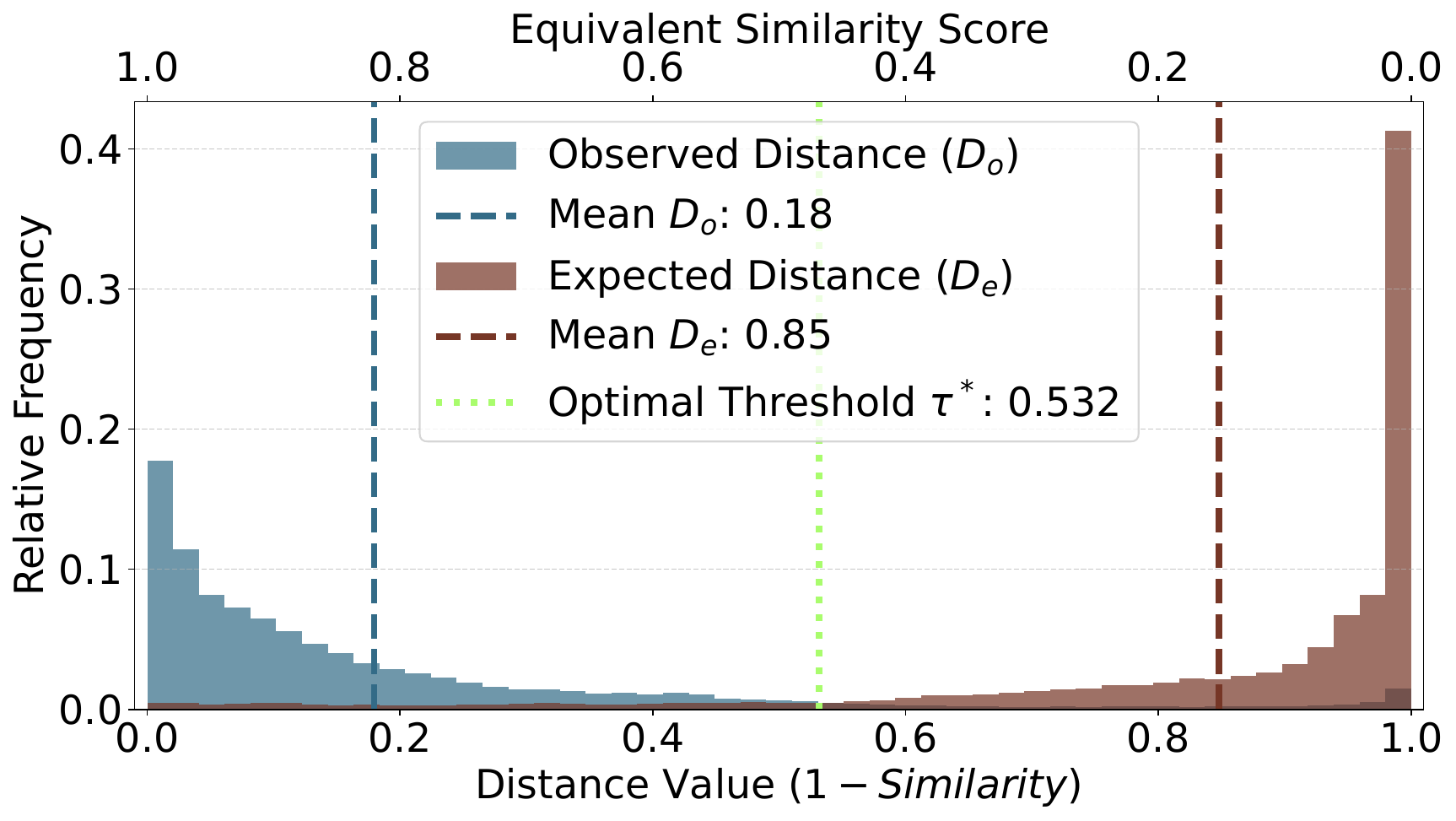}
    \caption{Distribution of Observed ($D_o$) vs. Expected ($D_e$) disagreement for instance segmentation on TexBiG using $1-\text{IoU}$.}
    \label{fig:do_de_texbig}
\end{figure}

We identify the optimal metric to measure the annotator intention by maximizing the statistical separation between observed disagreement ($D_o$) and expected chance disagreement ($D_e$) as shown for IoU in \cref{fig:do_de_texbig}. We quantify this separation using the Kolmogorov-Smirnov (KS) statistic. \cref{tab:KS_scores_texbig} presents the results for the TexBiG dataset.

\begin{table}[h]
    \centering
    \small
    \begin{tabular}{lcc}
        \toprule
        \textbf{Distance Metric} & \textbf{KS Statistic} $\uparrow$ & \textbf{Calibration Anchor $\tau^*$}\\
        \midrule
        Polygon IoU & \textbf{0.82} & $0.53$\\
        L2 Centroid & 0.78 & $0.02$\\
        Mask GIoU & 0.78 & $0.28$\\
        \bottomrule
    \end{tabular}
    \caption{TexBiG results for principled configuration.}
    \label{tab:KS_scores_texbig}
\end{table}

The data indicates that Polygon IoU achieves the highest separation (KS=0.8189), confirming it as the most robust metric for distinguishing annotator consensus from random chance in this domain. While the statistically optimal calibration anchor is 
 $\tau^*=0.532$, we adopt the standard threshold $\tau=0.5$. We select this value because it lies within the statistical margin of error (See robustness in \cref{sec:robustness_analysis}) of $\tau^*$ and maintains alignment with standard computer vision benchmarks (\eg, mAP@50), facilitating cognitive comparison for future research.

\textbf{Specification and dataset-wide mean Agreement.} Based on this calibration, we define the complete configuration as: 
\begin{equation} 
    \mathrm{K}\alpha\mathrm{LOS}{\big(d_{\text{IoU}},\tau{=}0.5,\mathcal{S}{=}\text{Greedy},\psi_{\text{soft}}\big)} 
\end{equation} 
    
Executing this configuration on the TexBiG dataset yields a dataset-wide mean \KripA{} of \textbf{0.9055}. This score indicates ``almost perfect'' agreement, but a scalar summary obscures the specific sources of disagreement. By standardizing the complex instance segmentation task into a nominal reliability matrix, the framework unlocks granular diagnostics regarding data ambiguity and annotator behavior.

\begin{figure}[t]
    \centering
    \includegraphics[width=\linewidth]{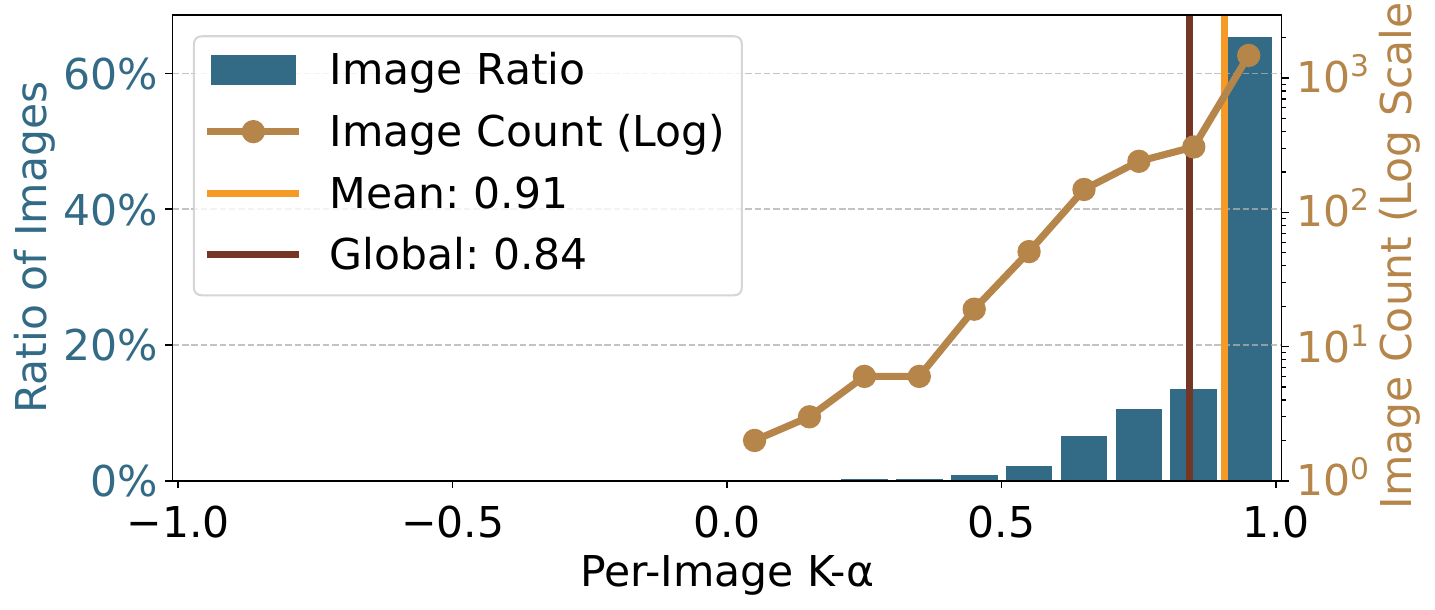}
    \caption{\textbf{Per-Image Agreement.} The left-skewed distribution reveals that while the median image achieves perfect consensus ($\alpha=1.0$), a long tail of disagreement exists, identifying specific ambiguous samples for review.}
    \label{fig:sub_distribution}
\end{figure}

\begin{figure}[t]
    \centering
    \includegraphics[width=\linewidth]{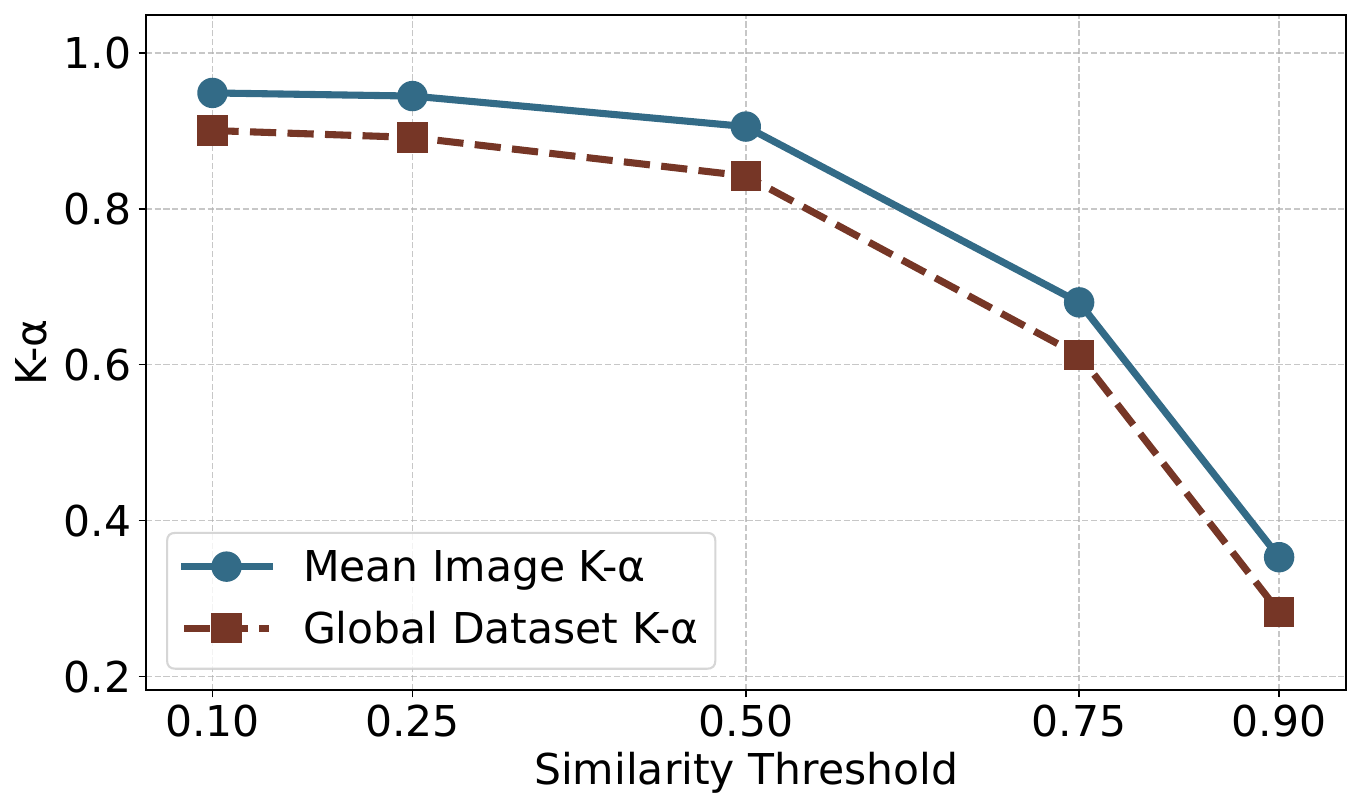}
    \caption{\textbf{Localization Sensitivity Analysis (LSA).} Agreement remains stable ($\alpha \approx 0.94 \to 0.90$) up to $\tau_s=0.5$ but collapses at $\tau_s=0.9$ ($\alpha=0.35$). This ``Cliff of Precision'' quantifies the limit of human spatial consistency.}
    \label{fig:sub_sensitivity}
\end{figure}

\begin{figure}[t]
    \centering
    \includegraphics[width=\linewidth]{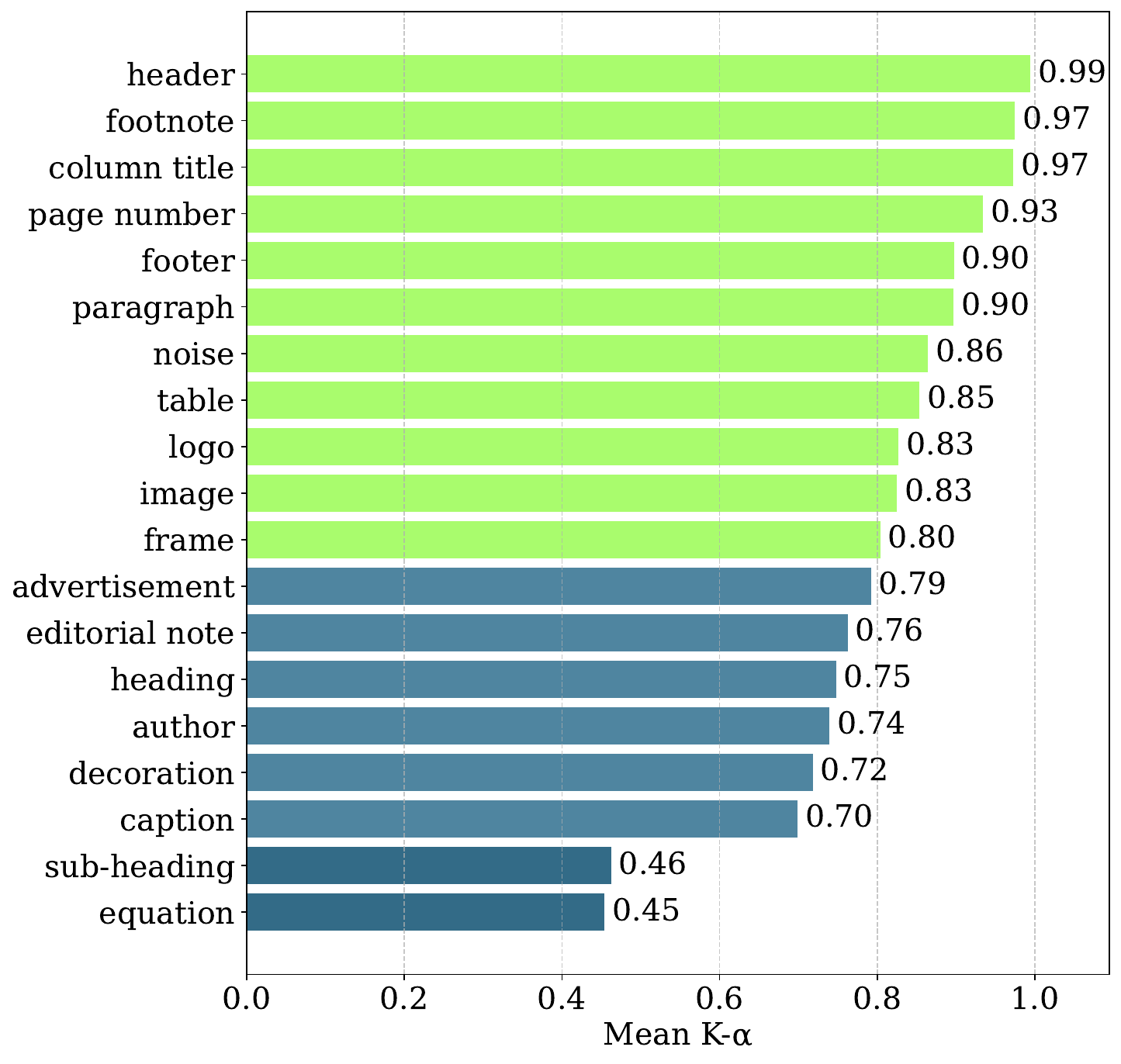}
    \caption{\textbf{Class Recognition Difficulty.} Structural elements like \texttt{header} ($\alpha=0.99$) are robust, whereas semantic definitions for \texttt{equation} ($\alpha=0.45$) and \texttt{sub-heading} ($\alpha=0.46$) require guideline refinement.}
    \label{fig:sub_difficulty}
\end{figure}

\textbf{Data Diagnostics.} The distribution analysis in \cref{fig:sub_distribution} confirms that the consensus is stable, with a median $\alpha=1.0$. However, the Localization Sensitivity Analysis (LSA) in \cref{fig:sub_sensitivity} exposes the physical limits of this consensus. Agreement remains high ($>0.90$) for thresholds $\tau_s \in [0.1, 0.5]$, indicating raters agree on object existence. The sharp drop at $\tau_s=0.9$ ($\alpha=0.35$) proves that pixel-perfect boundary consensus is unachievable in this domain. Furthermore, \cref{fig:sub_difficulty} isolates semantic ambiguity. While structural classes like \texttt{header} ($\alpha=0.99$) are solved, abstract concepts like \texttt{equation} ($\alpha=0.45$) generate significant noise, pinpointing exactly where the annotation guidelines fail.

\begin{figure}[t]
    \centering
    \includegraphics[width=1.0\linewidth]{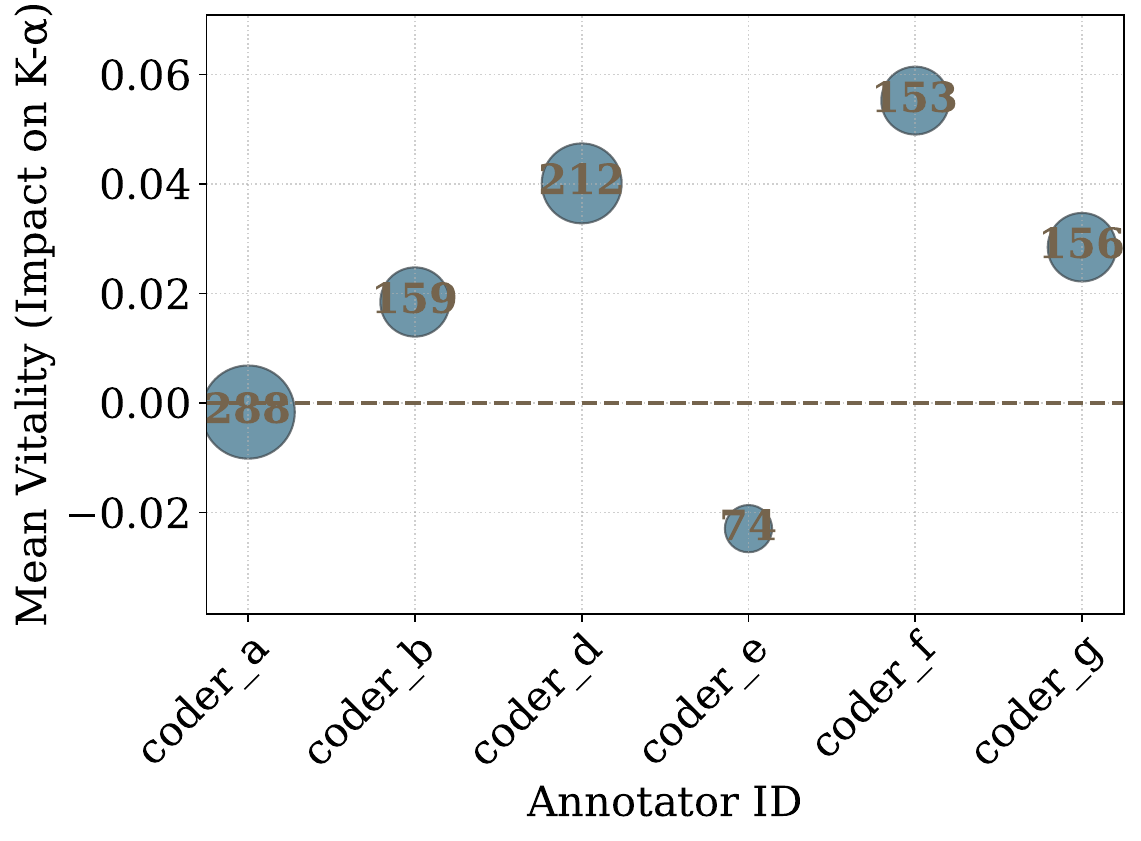}
    \caption{\textbf{Annotator Vitality.} \texttt{coder\_d} ($V_r=+0.04$) acts as a strong consensus builder, while \texttt{coder\_e} ($V_r=-0.02$) introduces systematic noise, suggesting a need for retraining.}
    \label{fig:sub_vitality}
\end{figure}

\begin{figure}[t]
    \centering
    \includegraphics[width=1.0\linewidth]{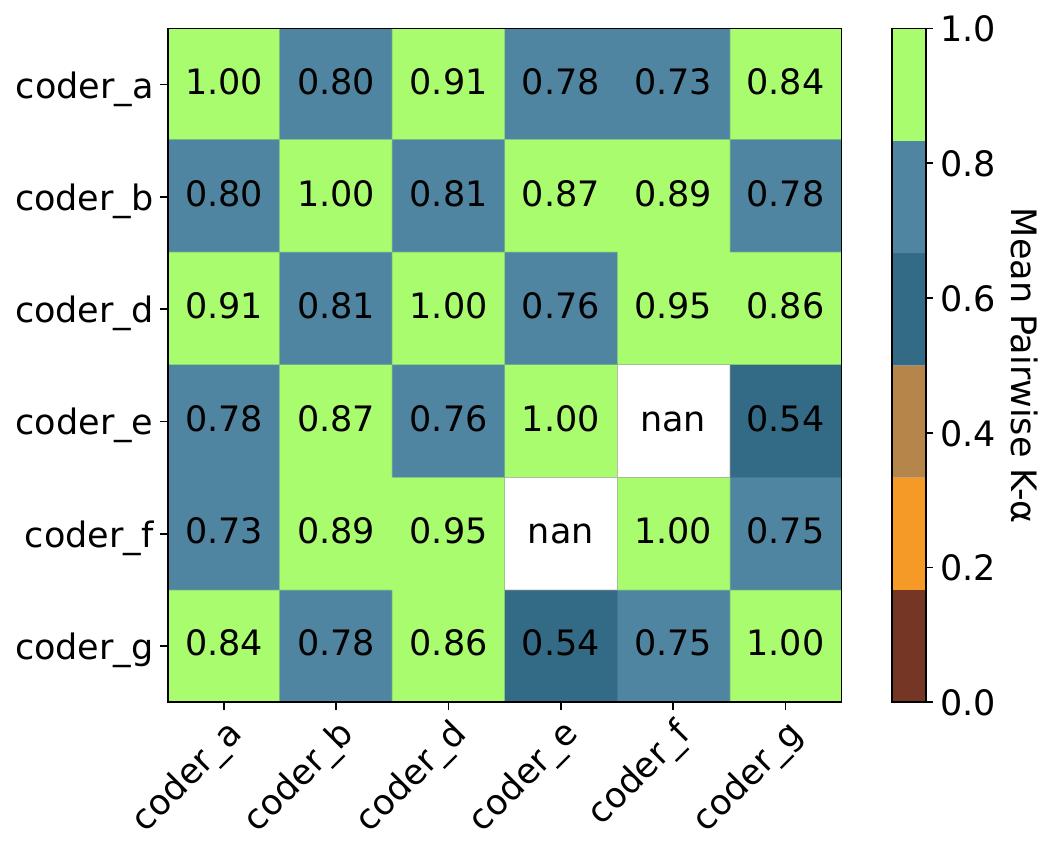}
    \caption{\textbf{Collaboration Clusters.} The heatmap reveals distinct ``schools of thought.'' \texttt{coder\_f} and \texttt{coder\_d} achieve higher pairwise agreement ($0.95$) than the global average, indicating shared implicit conventions.}
    \label{fig:sub_heatmap}
\end{figure}

\textbf{Annotator Diagnostics.} The framework isolates human factors without requiring ground truth. Annotator Vitality (\cref{fig:sub_vitality}) identifies \texttt{coder\_d} as a stabilizer who increases global agreement by $0.04$, whereas \texttt{coder\_e} is a source of divergence ($V_r = -0.02$). Finally, the Collaboration Heatmap (\cref{fig:sub_heatmap}) detects implicit ``schools of thought.'' The high pairwise agreement between \texttt{coder\_f} and \texttt{coder\_d} ($0.95$) versus \texttt{coder\_e} ($0.76$) suggests divergent interpretations of the task, allowing for targeted intervention rather than dataset pruning.

\subsection{3D Volumetric Instance Segmentation}
\label{sec:3D}

To demonstrate the transferability of \KalphaLOS{} from 2D images to 3D volumetric data, we apply the meta-algorithm to the LIDC-IDRI dataset \cite{armato2015lidc}. This task involves the binary classification (Nodule vs. Background) and localization of pulmonary nodules in CT scans. We restrict the analysis to nodules $\ge 3$ mm, as smaller instances lack explicit contour definitions in the source data. Furthermore, because LIDC-IDRI anonymizes annotator identities, we treat the four reading sessions per scan as independent, unidentifiable raters.

\begin{figure}[t]
    \centering
    \includegraphics[width=1.0\linewidth]{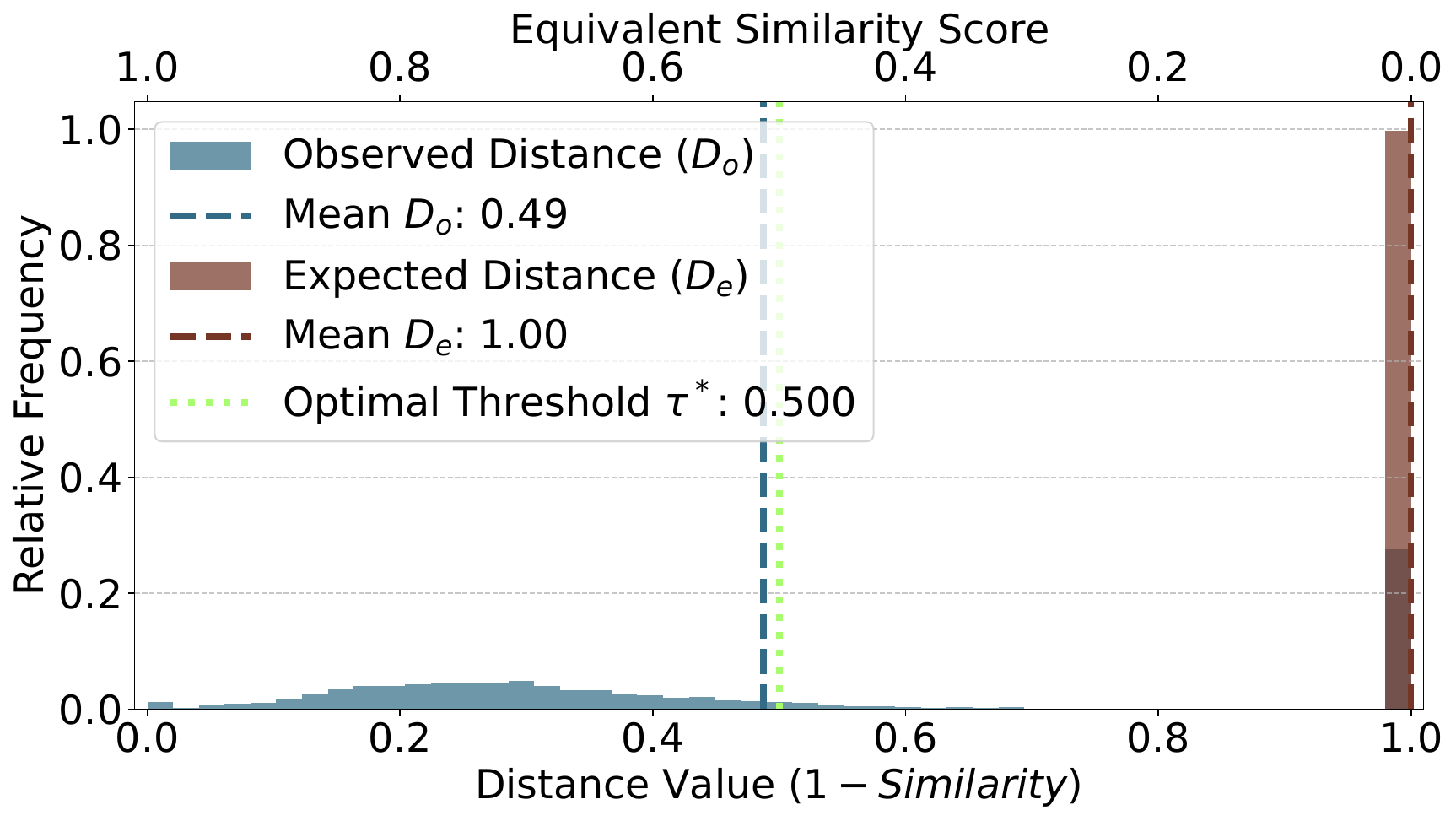}
    \caption{\textbf{Signal vs. Chance in 3D.} The distribution of Expected Disagreement ($D_e$) converges to a Dirac delta at $1.0$. This reflects the sparsity of nodules within the vast 3D volume; unlike in 2D images, the probability of two random annotations overlapping in a voxel grid is statistically negligible.}
    \label{fig:lidc_dode}
\end{figure}

\textbf{Principled Configuration.}
We adapt the distance metric to operate on voxel grids rather than pixel masks. Let $V_{ik}^r$ represent the set of voxels occupied by annotation $k$ from rater $r$ in volume $i$. We define the Volumetric IoU distance $d_{\text{vol}}$ as:
    \begin{equation}
    d_{\text{vol}}(V_{ik}^{a}, V_{il}^{b}) = 1 - \frac{|V_{ik}^{a} \cap V_{il}^{b}|}{|V_{ik}^{a} \cup V_{il}^{b}|}.
    \end{equation}
To determine the validity of this metric, we analyze the statistical separation between the Observed Disagreement ($D_o$) and Expected Disagreement ($D_e$). As shown in \cref{fig:lidc_dode}, the Kolmogorov-Smirnov (KS) statistic maximizes at a threshold of $\tau^*=0.50$ (KS $= 0.7237$). While one could search for alternative distance functions to further maximize this separation, the high KS score confirms that $d_{\text{vol}}$ successfully distinguishes genuine inter-rater correspondence from chance. Consequently, we settle on this metric and update the calibration anchor to the data-driven optimum:
\begin{equation}
\mathrm{K}\alpha\mathrm{LOS}_{\big(d_{\text{vol}},\,\tau{=}0.50,\,\mathcal{S}{=}\text{Greedy},\,\psi_{\text{soft}}\big)}.
\end{equation}

\textbf{Domain-Specific Topology.}
The distribution analysis of $D_{o}$ vs $D_{e}$ in \cref{fig:lidc_dode} reveals a unique characteristic of volumetric analysis: the Expected Disagreement ($D_e$) converges strongly to $1.0$. In 2D tasks, random boxes often overlap slightly, creating a "soft" chance distribution. In 3D, the minute spatial footprint of a nodule relative to the CT volume makes random overlap impossible. This simplifies the problem topology: since chance is effectively constant ($D_e \approx 1$), the reliability of the dataset depends entirely on the precision of the Observed Disagreement ($D_o$).

\begin{figure}[t]
    \centering
    \includegraphics[width=1.0\linewidth]{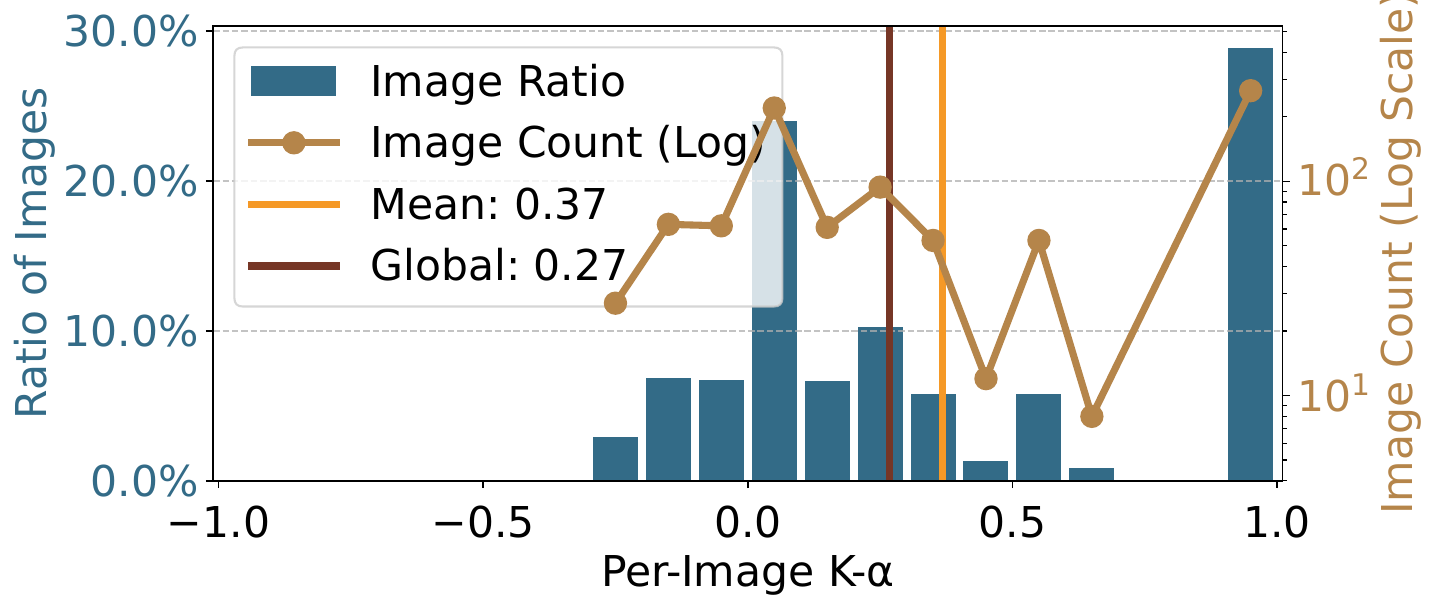}
    \caption{\textbf{Per-Image Agreement Distribution.} The distribution shows a significant spread (Mean $\alpha=0.37$). While a subset of scans achieves perfect consensus ($\alpha=1.0$), the heavy tail indicates that medical annotation involves inherent ambiguity not present in structural tasks like document layout.}
    \label{fig:lidc_distribution}
\end{figure}

\textbf{Results and Interpretation.}
The pipeline yields a dataset-wide mean agreement of $\alpha = 0.3683$. While significantly lower than the "Almost Perfect" scores observed in TexBiG ($0.9055$), this does not imply poor data quality. Rather, it quantifies the inherent ambiguity of the medical domain. As seen in the Per-Image Distribution (\cref{fig:lidc_distribution}), agreement varies wildly, with a heavy tail of difficult cases where experts fundamentally disagree on tissue characterization.

\begin{figure}[t]
    \centering
    \includegraphics[width=1.0\linewidth]{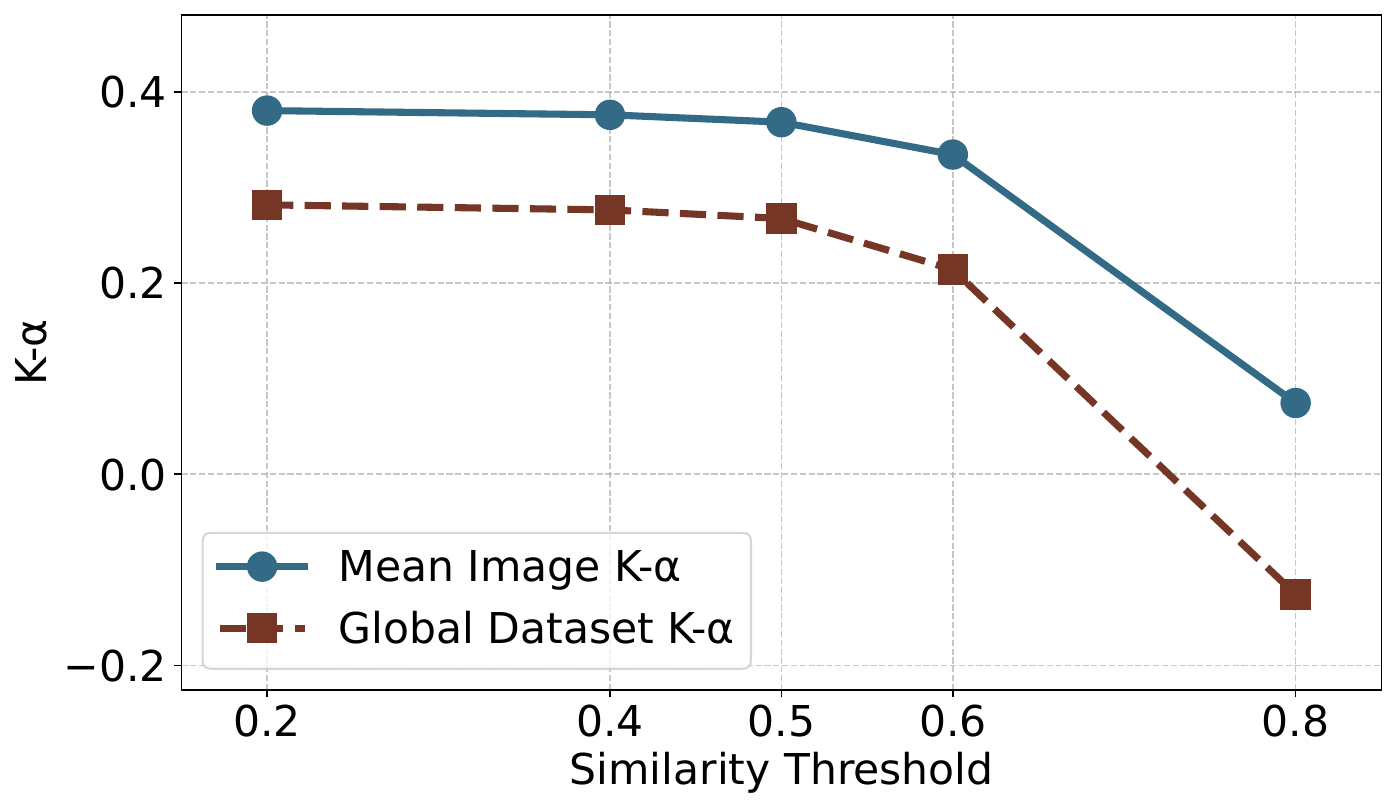}
    \caption{\textbf{Localization Sensitivity (LSA).} The agreement exhibits a ``Plateau'' behavior, remaining stable ($\alpha \approx 0.37$) up to $\tau_s=0.50$. Agreement starts to drop when leaving the calibration anchors confidence interval and drops significantly at $\tau_s=0.8$.}
    \label{fig:lidc_lsa}
\end{figure}

\textbf{Diagnostics.}
The Localization Sensitivity Analysis (LSA) in \cref{fig:lidc_lsa} clarifies the nature of this disagreement. The curve exhibits a ``Plateau'' stability from $\tau_s=0.2$ to $\tau_s=0.5$ ($\alpha \approx 0.37$), indicating that radiologists consistently agree on the \textit{anatomical identity} and rough location of nodules. However, the ``Cliff'' at $\tau_s=0.6$ (mean $\alpha \to 0.33$) reveals that they disagree on the \textit{morphological extent}. Unlike the sharp vectors of a document layout, biological tissue lacks discrete boundaries, making pixel-perfect segmentation consensus ($\tau_s > 0.6$) achievable only for the most obvious instances.

\subsection{Pose Estimation}
\label{sec:pose_estimation}

Finally, we extend \KalphaLOS{} to the domain of articulated pose estimation using the MARS dataset~\cite{segalin2021}. This dataset captures the social interactions of mice via a fixed overhead or frontal camera in a standardized cage environment. Unlike bounding boxes or segmentation masks, pose annotations consist of a structured graph of keypoints (\eg, nose, ears, hips). This requires a distance metric that captures both the internal structural consistency (pose) and the global placement (location), while being invariant to image resolution.

\textbf{Distance Function.}
We use an adapted Image-Normalized Mean Per-Joint Position Error (N-MPJPE). Let $K^A = \{k_1^A, \dots, k_M^A\}$ be the set of $M$ keypoints provided by rater A, with coordinates normalized to the image dimensions. The distance function is defined as the average error over the union of visible keypoints $J = V_A \cup V_B$:
\begin{equation}
d_{\text{pose}}(K^A, K^B) = 1 - \underbrace{\frac{1}{|J|} \sum_{j \in J} \delta(k^A_j, k^B_j)}_{\text{N-MPJPE}}
\end{equation}
where the per-joint error $\delta$ accounts for both spatial deviation and visibility disagreement:
\begin{equation}
\delta(k^A_j, k^B_j) =
\begin{cases}
\frac{\| k^A_j - k^B_j \|_2}{\sqrt{2}} & \text{if } j \in V_A \cap V_B \\
1.0 & \text{if } j \in V_A \oplus V_B \\
\end{cases}
\end{equation}
For keypoints visible to both raters, we calculate the Euclidean distance normalized by the image diagonal ($\sqrt{2}$ in relative coordinates). For keypoints where visibility is disputed (present in one but not the other), we assign a fixed penalty of $1.0$.

\begin{figure}[t]
    \centering
    \includegraphics[width=1.0\linewidth]{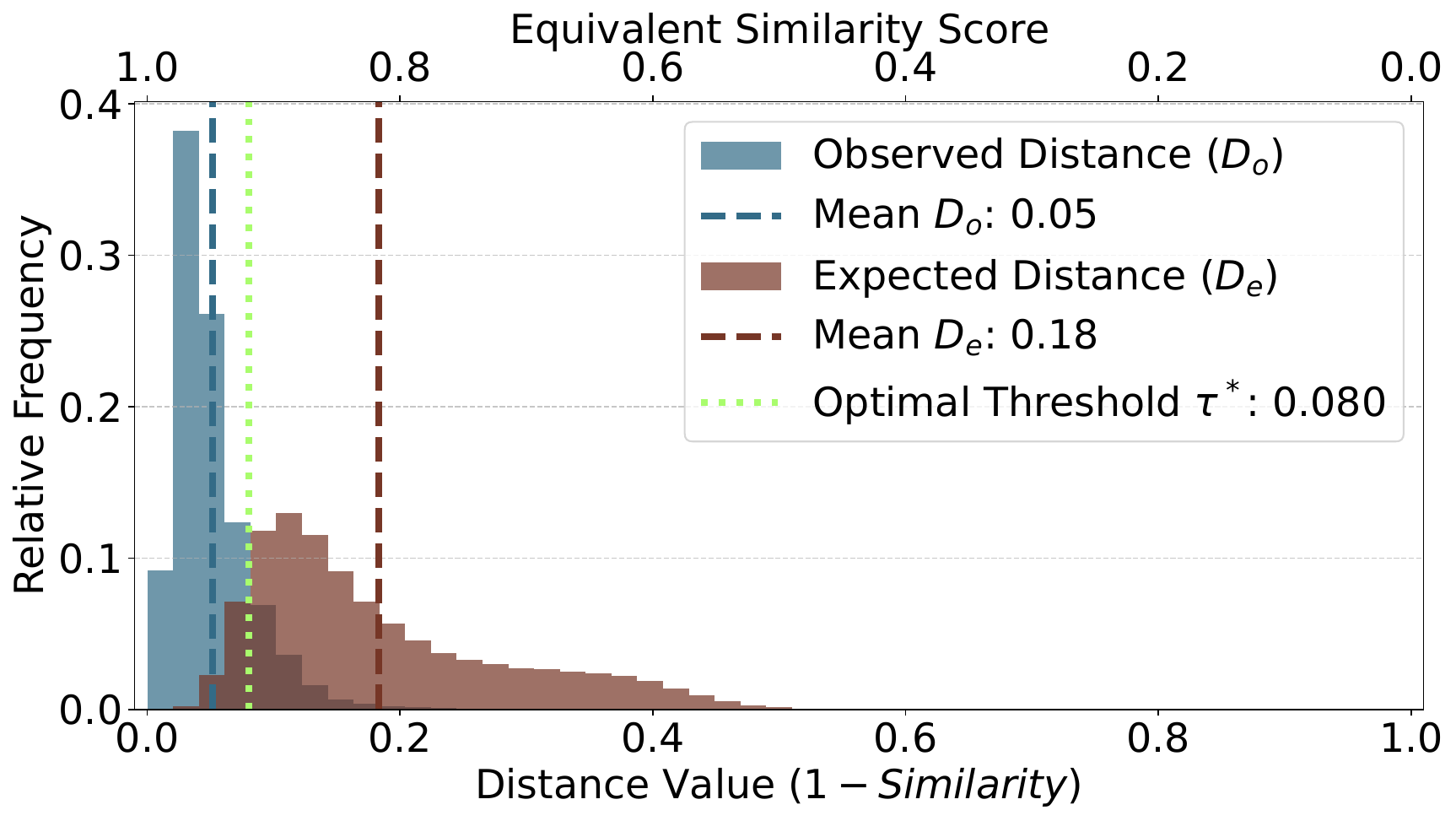}
    \caption{\textbf{High Noise Floor.} The distribution of Expected Disagreement ($D_e$) centers at a remarkably low distance of $0.18$. Due to the constrained cage environment and fixed camera, a random mouse pose spatially resembles the target pose, necessitating a strict threshold ($\tau_s^*=0.92$ or as distance $\tau^*=0.08$) to statistically isolate the annotator signal ($D_o$).}
    \label{fig:do_de_mars}
\end{figure}

\textbf{Threshold Adaptation and Validation.}
Validating $d_{\text{pose}}$ on MARS reveals a critical insight enabled by our framework (\cref{fig:do_de_mars}). The distributions of Observed Disagreement ($D_o$) and Expected Disagreement ($D_e$) are remarkably close, with means of $0.05$ and $0.18$ respectively. This proximity reflects the highly constrained nature of the dataset: since the mouse is filmed from a fixed overhead camera in a standard cage~\cite{segalin2021}, a randomly selected mouse (Chance) is spatially likely to be very close to the target mouse.
This narrow margin necessitates a principled selection of the matching threshold. The Kolmogorov-Smirnov (KS) statistic maximizes at a distance of $0.08$ ($d_{pose}$), corresponding to a similarity threshold of $\tau_s^*=0.92$ for N-MPJPE (KS $= 0.7630$). This data-driven calibration confirms that a strict similarity threshold is not merely a preference for precision, but a statistical requirement to separate true matches from environmental coincidence.

\begin{figure}[t]
    \centering
    \includegraphics[width=1.0\linewidth]{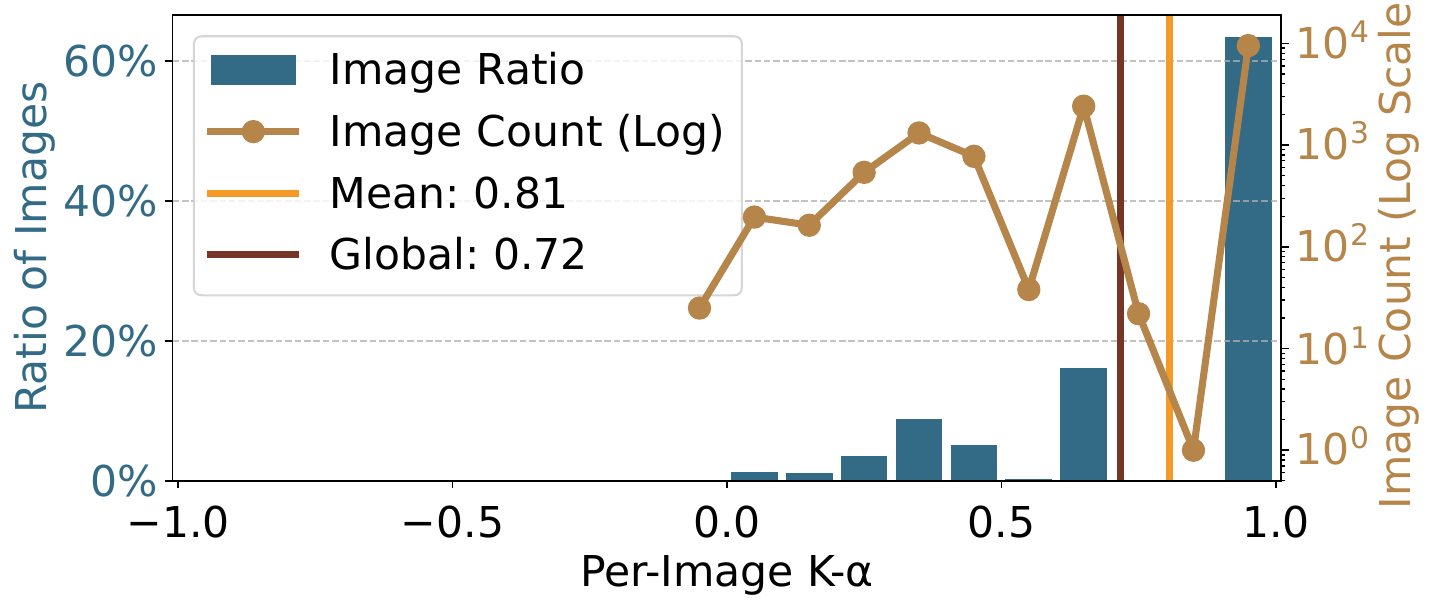}
    \caption{\textbf{Per-Image Agreement.} The high median ($\alpha=1.0$) indicates that for most frames, annotators achieve perfect consensus. However, the distribution tail reveals specific frames where occlusion or motion blur degrades reliability.}
    \label{fig:mars_alpha_dist}
\end{figure}

\textbf{Results and Analysis.}
Using the configuration $\mathrm{K}\alpha\mathrm{LOS}_{\big(d_{\text{pose}},\,\tau{=}0.08,\,\mathcal{S}{=}\text{Greedy},\,\psi_{\text{soft}}\big)}$, we obtain a dataset wide mean agreement of $\alpha = 0.8069$. To unpack this result, we examine the distribution of agreement scores across individual images (\cref{fig:mars_alpha_dist}). The distribution reveals a significant spread with a median of $1.00$. The high median indicates that for the majority of frames, annotators achieve near-perfect consensus on pose. However, the tail of the distribution highlights a subset of challenging frames where agreement degrades.

\begin{figure}[t]
    \centering
    \includegraphics[width=1.0\linewidth]{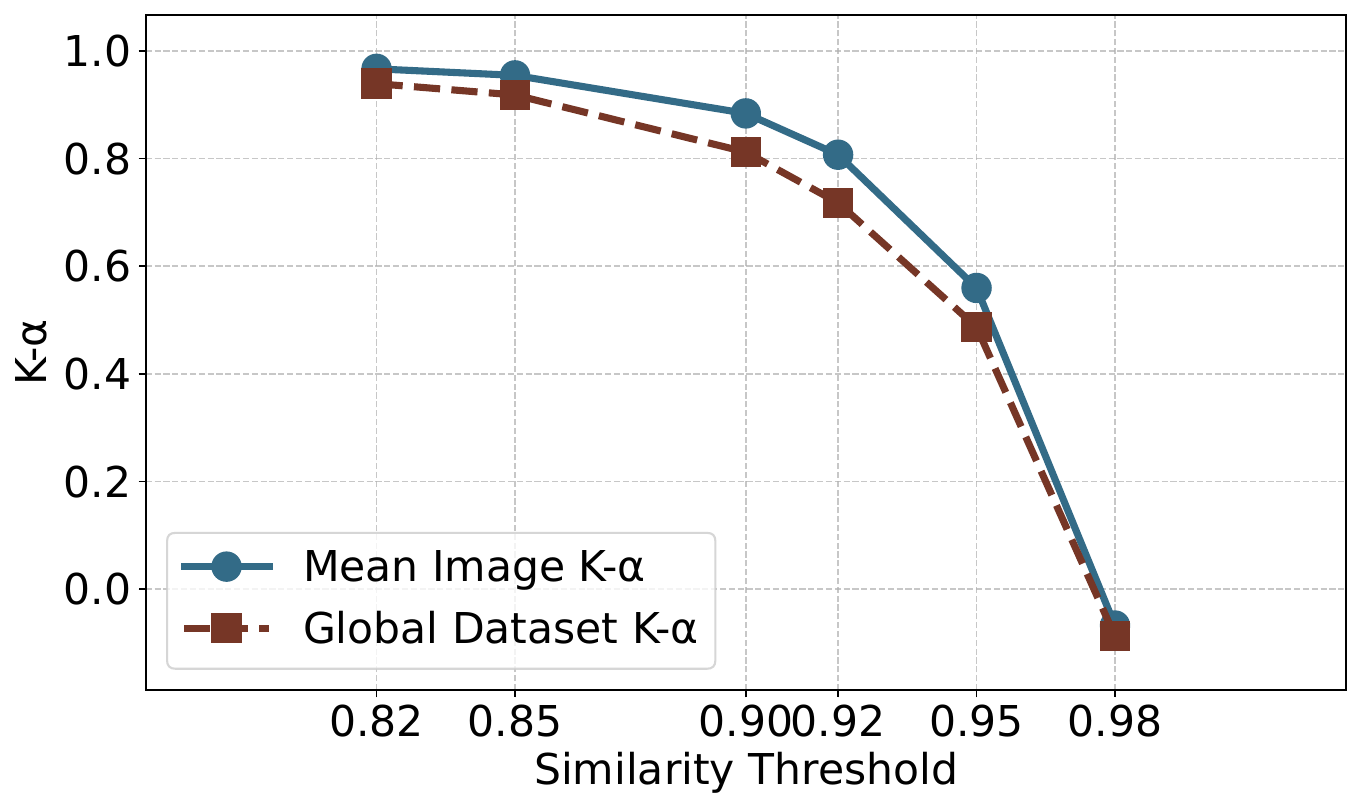}
    \caption{\textbf{Localization Sensitivity (LSA).} The metric remains stable across standard thresholds but drops at $\tau_s=0.95$, revealing the upper bound of inter-annotator precision.}
    \label{fig:mars_sensitivity}
\end{figure}

\textbf{Diagnostics.}
The Localization Sensitivity analysis (\cref{fig:mars_sensitivity}) confirms the logical relationship between strictness and agreement. As the similarity threshold tightens from $\tau_s=0.82$ to the calibrated anchor $\tau_s^*=0.92$, $\alpha$ remains relatively stable ($0.97 \to 0.81$), indicating robust agreement on the general pose structure. However, at the strict threshold of $\tau_s=0.98$, agreement drops sharply to $-0.07$. This inflection point identifies the limit of human precision: while raters agree on the overall pose, sub-pixel precision at the $0.98$ similarity level is not achievable even in this constrained domain.